%% file: main.tex
\documentclass[journal,twoside,web]{ieeecolor}

\def\BibTeX{{\rm B\kern-.05em{\sc i\kern-.025em b}\kern-.08em
    T\kern-.1667em\lower.7ex\hbox{E}\kern-.125emX}}

\input{macros}

\begin{document}

\title{Distributed Optimization via Kernelized Multi-armed Bandits}
\author{Ayush Rai and Shaoshuai Mou 
\thanks{This paragraph of the first footnote will contain the date on 
which you submitted your paper for review.}
\thanks{ Ayush Rai and Shaoshuai Mou are with the School of Aeronautics and Astronautics, Purdue University, West Lafayette, IN 47906 USA (e-mail: rai29@purdue.edu; mous@purdue.edu).}}

\maketitle

\begin{abstract}
Multi-armed bandit algorithms provide solutions for sequential decision-making where learning takes place by interacting with the environment. In this work, we model a distributed optimization problem as a multi-agent kernelized multi-armed bandit problem with a heterogeneous reward setting. In this setup, the agents collaboratively aim to maximize a global objective function which is an average of local objective functions. The agents can access only bandit feedback (noisy reward) obtained from the associated unknown local function with a small norm in reproducing kernel Hilbert space (RKHS). We present a fully decentralized algorithm, Multi-agent IGP-UCB (MA-IGP-UCB), which achieves a sub-linear regret bound for popular classes for kernels while preserving privacy. It does not necessitate the agents to share their actions, rewards, or estimates of their local function. In the proposed approach, the agents sample their individual local functions in a way that benefits the whole network by utilizing a running consensus to estimate the upper confidence bound on the global function. Furthermore, we propose an extension, Multi-agent Delayed IGP-UCB (MAD-IGP-UCB) algorithm, which reduces the dependence of the regret bound on the number of agents in the network. It provides improved performance by utilizing a delay in the estimation update step at the cost of more communication.
\end{abstract}



\section{Introduction}
\label{sec:introduction}
\IEEEPARstart{T}{he} problem of distributed optimization deals with the optimization of a function over a network of agents in which the whole function is not completely known to any single agent \cite{yang2019survey, zheng2022review}. In fact, the "global" function can be expressed as an average of "local" functions associated with each agent which are independent of one another. In particular, our interest lies in the case when these local functions are non-convex, unknown, and expensive to compute or record. To form a feasible problem, we assume that these local functions belong to a reproducing kernel Hilbert space (RKHS), which is a very common assumption in the literature \cite{vakili2021optimal, janz2020bandit, srinivas2009gaussian}. When dealing with unknown functions, the problem for each agent can be broken down into two segments: \textit{sampling} and \textit{optimization}. The objective holds to optimize (in our case maximize) the global function with as less number of samples as possible. The approach intuitively warrants some sort of cooperation and communication among agents, as the function each agent samples from is different from the function it is aiming to optimize. Numerous examples and diverse practical applications motivate our problem setup. These include but are not limited to the field of distributed machine learning \cite{cano2016towards, anandkumar2011distributed, mcmahan2017communication}, wireless sensor networks \cite{tran2012long}, distributed model fitting \cite{boyd2011distributed}, and multi-agent control systems \cite{nedic2018distributed}.

The sequential decision-making of our problem setup makes a very close resemblance with cooperative multi-agent multi-armed bandits \cite{kanagawa2018gaussian}. In a classical single-agent multi-armed bandit, an agent has access to $K$ arms. At each time step the agent pulls an arm and in return receives a stochastic reward sampled from an unknown distribution with an unknown mean which it seeks to maximize \cite{auer2002finite, agrawal1995continuum}. The performance of the agent is measured in terms of cumulative regret, where the regret is the reward lost had the distribution known  a priori. For now, if we consider a general problem of function optimization (where the function belongs to RKHS) in discrete space (say $\mathcal{D}$), then it can be seen as a multi-armed bandit problem with $|\mathcal{D}|$ arms. The peculiarity in the optimization problem is that distributions of the rewards associated with the arms are related to one another via a kernel function. These are often referred to as kernelized bandit or bandit optimization problems. 

The multi-agent setting of kernelized bandit has been explored in many flavors. These problems have mainly concentrated on enhancing the learning of individual agents (agent’s local function) by capitalizing on similarity between function \cite{dubey2020cooperative,li2022communication,du2021collaborative} or delved into cooperative learning of global function from federated aspects to address privacy concerns, albeit not in a decentralized manner for arbitrary networks \cite{salgia2023collaborative,dai2020federated_bay}. To the best of the authors’ knowledge, this work is the first to propose a fully distributed strategy for kernelized bandit problem with heterogeneous reward setting which can be implemented on an arbitrary network. This allows us to solve the classical distributed optimization problem for unknown and non-convex functions which hasn’t been done before.

\subsection{Related work}

\textit{a) Distributed Optimization: }
Distributed optimization has been an active area of research, primarily focusing on convex or strongly convex functions \cite{yang2019survey, zheng2022review}. Classical approaches for unconstrained distributed optimization, include distributed subgradient methods \cite{nedic2009distributed, lobel2010distributed}, the alternating direction method of multipliers \cite{boyd2011distributed, aybat2017distributed}, distributed proportional-integral (PI) control algorithm \cite{gharesifard2013distributed, kia2015distributed}, the fast gradient method \cite{jakovetic2014fast}, the push-sum method \cite{xi2018linear}, and proximal algorithms \cite{dhingra2018proximal, aybat2015asynchronous}. Researchers have also explored non-convex distributed optimization problems, such as in \cite{tatarenko2017non}, where the perturbed push-sum algorithm was generalized from convex to non-convex cases. The authors guarantee almost sure convergence to a local minimum of the global objective function if no saddle points exist. Additionally, gradient-free approaches for distributed optimization with unknown cost functions have been presented in \cite{pang2019randomized, yuan2014randomized}, assuming convex local functions. In contrast to this existing literature, our work proposes a novel sampling-based distributed algorithm for global optimization that considers unknown and non-convex functions. 

\textit{b) Cooperative Bandits:}
A wide range of approaches have explored the multi-armed bandit (MAB) problem in multi-agent network settings \cite{shahrampour2017, landgren2016distributed, landgren2016distributed2, moradipari2022collaborative, chakraborty2017coordinated, korda2016distributed, szorenyi2013gossip, martinez2019decentralized, scaman2017optimal, zhu2023distributed}. In a homogeneous reward setting, each agent faces the same MAB problem, and the network cooperatively estimates the arm with the largest expected reward. The work in \cite{korda2016distributed} studies a setup where, at each time step, an agent only communicates with one other agent chosen randomly. In \cite{chakraborty2017coordinated}, the authors consider the case in which agents can broadcast the last obtained reward to the network or take an action. The distributed bandits in \cite{landgren2016distributed, landgren2016distributed2, martinez2019decentralized} investigate the case where the group collaboratively shares their estimates of rewards and the number of times an arm was pulled over a fixed communication graph. Particularly in \cite{martinez2019decentralized}, the authors use delayed rewards and accelerated mixing to demonstrate improved numerical performance.

In a heterogeneous setting, it is considered that each arm has an agent-dependent reward distribution. The goal is to find the arm with the largest expected reward, averaged among agents. This setting was considered in \cite{shahrampour2017} with the setup that, at each round, all the agents collectively take the same action, which is decided by a majority vote. The work in \cite{moradipari2022collaborative} deals with linear bandit problems with a similar setup as in \cite{shahrampour2017}, but at each round, one action is randomly selected from the actions proposed by each agent and played as the network action. A competitive multi-agent bandit setup was considered in \cite{bistritz2018distributed, mehrabian2020practical} in which no reward is given to agents if they sample the same action. Heterogeneity is also explored in \cite{mitra2021exploiting}, where a federated algorithm building on a successive elimination technique is proposed, and the globally best arm is defined as the arm with the highest expected reward among the locally best arms. The degree of dissimilarity between the probability distributions of arms associated with different agents is used as a measure to decrease the required communication rounds. 
In \cite{shi2021per}, the authors proposed a personalized federated multi-armed bandit framework that balances the global reward (generalization) and local rewards (personalization). This work was further extended in \cite{reda2022near}, where the global reward of an arm is a weighted sum of the rewards for all agents. Our work expands upon the heterogeneous setting within the context of kernelized bandit problems, where an estimate of the distribution of one arm also provides some insight into the distributions of other arms.

\textit{c) Gaussian Process Bandit Optimization:} Bayesian optimization problems have been extensively studied in the literature, particularly employing reinforcement learning-based approaches \cite{brochu2010tutorial}. Within the bandit setting, Gaussian process optimization gained significant popularity with the introduction of the GP-UCB algorithm \cite{srinivas2009gaussian}, which established a sub-linear regret bound by leveraging information gain bounds. The IGP-UCB algorithm presented in \cite{chowdhury2017kernelized} further improved this regret bound by introducing a new self-normalized inequality initially developed in \cite{abbasi2011improved}. Another approach, the Kernel-UCB algorithm, proposed in \cite{valko2013finite}, provided an alternative regret bound based on the effective dimension of the data's projection in the RKHS. Contextual GP optimization was addressed using the CGP-UCB algorithm in \cite{krause2011contextual}, while time-varying GP optimization was introduced in \cite{bogunovic2016}.

In a multi-agent setting, \cite{dubey2020kernel} proposes the COOP-KernelUCB algorithm for a setup where agents cooperatively aim to optimize their own function. The cooperation part comes into the picture due to the relationship among the agents' reward functions, resulting in a decreased cumulative regret compared to independent learning. 
In \cite{du2021collaborative}, the authors propose a collaborative pure exploration in kernel bandit algorithm where agents solve different but related tasks. In \cite{li2022communication, lilearning}, the authors present a communication-efficient approach for kernelized bandits 
in a distributed setting where the same reward function is shared by all the agents. In all of these works, some similarity is assumed in the local reward functions. 
However, we consider the problem where the functions agents are \textit{sampling} from are independent of each other, while the function that the agents want to \textit{optimize} is the average of all the local reward functions.

\subsection{Contributions}
In this work, we solve the problem of distributed optimization over a connected network of agents by framing it as a kernelized multi-agent multi-armed bandit problem. To the best of our knowledge, this is the first attempt at solving the problem of multi-agent Gaussian process bandit optimization with an independent local reward setting. Our main contributions are summarized as follows:

\begin{itemize}
    \item Firstly, we present a novel problem formulation for distributed optimization that differs from previous literature in several important ways. Specifically, we consider a more generic problem, where the local functions corresponding to each agent are independent, unknown, and non-convex. We do assume the functions to have a low norm in a reproducing kernel Hilbert space (RKHS).
    \item Secondly, as a kernelized bandit problem, we consider a heterogeneous reward function setting. In contrast to previous works where individual agents aim to optimize their own functions (but related), we aim to maximize the global function in a case where the functions of individual agents are independent of each other, resulting in a problem that cannot be solved by any single agent. Our proposed regret metric, detailed in Section 2, captures the joint impact of each agent's decision on the overall network performance.
    \item Third, by utilizing the theory of distributed consensus and the IGP-UCB (Improved Gaussian process upper confidence bound) algorithm, we propose a cooperative MA-IGP-UCB algorithm for connected networks. We provide a sub-linear upper bound for popular classes of kernels\footnote{Squared exponential kernel and Matérn kernel with $\nu>d/2$.} on cumulative regret of $\mathcal{O}^{*}(N^2\sqrt{T}(B\sqrt{\gamma_T}+\gamma_T))$\footnote{The notation $\mathcal{O}^{*}$ is used to suppress the logarithmic factors.} for connected networks (see Theorem \ref{thm1}), and $\mathcal{O}^{*}(\sqrt{T}(B\sqrt{\gamma_T}+\gamma_T))$ for completely connected networks (see Corollary \ref{cor1}). Furthermore, we provide an extension, the MAD-IGP-UCB algorithm which runs in stages to reduce the dependence of regret bound on the number of agents.
    \item Finally, the proposed algorithm does not require the agents to share their estimate of their local function, actions taken, or rewards received. The agents only share their estimate of the global function with their neighbors, providing a privacy preserving algorithm. 
\end{itemize}

The article is divided into 8 sections. Section \ref{sec:prob_formulation} provides an overview of the problem formulation, explaining the problem of distributed optimization (framed as a maximization problem) and introducing the concepts of instantaneous and cumulative regret, which are crucial in understanding the solution to the problem. Section \ref{sec:preliminaries} offers background information on Gaussian process optimization and information bounds, laying the groundwork for our algorithms. The main algorithm (MA-IGP-UCB) is presented in Section \ref{sec:algorithm}, with the associated regret bounds discussed in Section \ref{sec:regret_bounds}. Section \ref{sec:extension} introduces an extension of the main algorithm (MAD-IGP-UCB), incorporating the concepts of delayed rewards and mixed estimates. The numerical performance of the algorithms is detailed in Section \ref{sec:experiments}. Finally, the conclusion and future directions are provided in Section \ref{sec:conclusion}.

\textit{Notations:} We use the notation $\mathbb{R}^n$ and $\mathbb{R}^{n\times p}$ to denote the set of all $n \times 1$ real vectors and $n \times p$ real matrices, respectively. The transpose of a matrix or vector is denoted by $(\cdot)^{\top}$. The absolute value of a scalar $v$ is denoted by $|v|$.   A column stack of vectors $x_i, i=1,2,\dots,r$ is written as  $\col\{x_1,\dots, x_r \}$. Here $\odot$ denotes the Hadamard (element-wise) product and $\mathbf{1}_r$ denotes a column vector in $\R^r$  where all its entries are equal to $1$.

\section{Problem Formulation}
\label{sec:prob_formulation}

The distributed optimization problem we target in this work is
\begin{align} \label{eq_opti0}
\max_{x\in \mathcal{D}} \quad  F(x) \triangleq \frac{1}{N}\sum_{i=1}^N f_i(x) 
\end{align}  
using a multi-agent system (MAS) of $N$ agents, whose communication network can be described by a fixed connected undirected graph $\mathcal{G} = (\mathcal{V},\mathcal{E})$, where $\mathcal{V} = \{1,...,N\}$ denote the node set and $\mathcal{E} \subseteq \mathcal{V}\times\mathcal{V}$ denote the edge set. The \textit{Neighbor} set of agent $i$ is denoted by $\mathcal{N}_i = \{ j \in \mathcal{V} | (j,i) \in \mathcal{E}\}$. The functions $F(x): \mathcal{D} \rightarrow \mathbb{R}$ and $f_i(x):\mathcal{D} \rightarrow\mathbb{R}$ are the \emph{global reward function} and \emph{local reward function} associated with agent $i$, respectively; where $\mathcal{D}  \subset \mathbb{R}^n $ is a compact, convex set, defining the action space for each agent. Instead of knowing the analytical form of $f_i$, we suppose agent $i$ can only access a noisy observation at time $t$:
\begin{align}
    y_{i,t}= f_i(x_{i,t}) + \epsilon_{i,t},
    \label{eq:noise_observation}
\end{align}
where $x_{i,t} \in \mathcal{D}$ denotes the action of agent $i$ taken at time $t$ and $\epsilon_{i,t}$ denotes an i.i.d. observation noise. The goal of distributed optimization is to find an iterative way for each agent $i$ to update $x_{i,t}$ only using information from its neighbors $\mathcal{N}_i$ such that $\lim_{t\rightarrow \infty} x_{i,t}= x^*, \ \forall i=1,2,...,N$, where $x^*$ solves the optimization problem in (\ref{eq_opti0}).

Since each agent may take a different action at any given time $t$, we measure the reward received by the network as the average of the global function evaluated at each agent's individual action, i.e., $\frac{1}{N}\sum_{j=1}^N F(x_{j,t})$. It is important to note that the maximum of this network reward is only achieved when all agents take the optimal action, i.e., $x_{j,t} = x^{*} \; \forall j=1,\dots,N$ with value $F(x^*)$.\footnote{In case of multiple optimal points, the agents' actions, $x_{j,t}$, can converge to any of them.} Thus, maximizing the sum of the network's reward over time, $\sum_{t=1}^{T}\frac{1}{N}\sum_{j=1}^N F(x_{j,t})$, can be seen as the objective of achieving \eqref{eq_opti0} as quickly as possible. In this context, cumulative regret naturally arises as a performance metric that measures the loss of rewards due to not knowing the point $x^{*}$ in advance. The \emph{cumulative regret} is defined as $R(T) = \sum_{t=1}^T r_t$, where $r_t$ denotes the \emph{instantaneous pseudo-regret}:
\begin{align}
    r_t &= \frac{1}{N}\sum_{i=1}^N f_i(x^{*}) - \frac{1}{N^2}\sum_{i=1}^N \sum_{j=1}^K f_i(x_t^j)  \nonumber \\
    &= F(x^{*}) -  \frac{1}{N}\sum_{j=1}^NF(x_t^j).
    \label{eq:reg_definition}
\end{align}
If the cumulative regret grows sub-linearly with $T$, then the ratio $R/T \rightarrow 0$ as $T$ approaches infinity, causing the instantaneous pseudo-regret to vanish, which maximizes the global reward $F$. Therefore, the distributed optimization for unknown functions problem can be solved using a distributed strategy that minimizes the cumulative regret $R(T)$.

\begin{remark}
The pseudo-regret capture the joint impact of all agents' decisions on the overall performance of the network. No individual agent can reduce the regret without communication. This notion of regret is in close spirit with what is commonly used in the literature dealing with multi-agent bandit problems with heterogeneous rewards \cite{zhu2023distributed, shi2021federated,shahrampour2017}.
\end{remark}

One cannot generally solve the problem for arbitrary reward functions, even in a single-agent case \cite{chowdhury2017kernelized, bogunovic2016}. To make the problem well-defined, we impose certain assumptions on the observation noises and reward function, which are standard in the bandit literature \cite{chowdhury2017kernelized, srinivas2009gaussian,dubey2020cooperative}.

\begin{assumption} For each agent $i$, the noise sequence to be $\{\epsilon_{i,t}\}_{t=1}^{\infty}$ is assumed to be conditionally sub-Gaussian for a fixed known constant $\eta\geq 0$, i.e.
\begin{align}
    \mathbb{E}[ e^{\lambda\epsilon_{i,t}} | \mathcal{F}_{t-1}] \leq \exp\left(\frac{\lambda^2 \eta^2}{2}\right) \quad \forall \lambda \in \mathbb{R},\quad \; \forall t\geq 1,
\end{align}
where $\mathcal{F}_{t-1}$ is the $\sigma-$algebra generated by the random variable $\{x_{i,\tau},\epsilon_{i,\tau}\}_{\tau=1}^{t-1}$ and $x_{i,t}$. 
\label{assumption1}
\end{assumption}

\begin{assumption} For each agent $i$, the unknown function $f_i$ is assumed to have a small norm in the RKHS, with positive semi-definite kernel function $k:\mathcal{D \times D \rightarrow \R}.$ We assume that $f_i$'s have a known bound, $B$, on the RKHS norm, $\|f_i\|_k = \sqrt{\braket{f,f}_k}$, which provides a bound on the smoothness of $f_i$ with respect to the kernel function $k$ \cite{chowdhury2017kernelized}. Furthermore, we also assume bounded variance by restricting $k(x,x) \leq 1 \;, \forall x\in \mathcal{D}$.
\label{assumption2}
\end{assumption}

\section{Preliminaries: Gaussian process optimization and Information Bound}
\label{sec:preliminaries}

We first review the idea of GP bandit algorithms for single agent \cite{srinivas2009gaussian,chowdhury2017kernelized,valko2013finite}. These algorithms start with an initial Gaussian process prior distribution of $GP_D(0,k(x,x))$ for the unknown reward function $f(x), \forall x\in \mathcal{D}$, where $k(x,x)$ is the kernel function associated with RKHS $H_k(\mathcal{D})$. As $\epsilon_t$ are assumed to be sub-Gaussian, they can be drawn independently from $\mathcal{N}(0,\lambda)$, with $\lambda>0$. If the set of sampling points is given by $\A_{1:t} = [x_1, \dots, x_t]$, then it can be assumed the observed rewards $y_{1:t}=[y_1,\dots,y_t]$ have the multivariate Gaussian distribution $\mathcal{N}(0,(K_t+\lambda I))$, where $K_t = [k(x,x')]_{x,x'\in \A_{1:t}}$. By utilising the fact that $y_{1:t}$ and $f(x)$ are jointly Gaussian, the posterior distribution over $f$, condition on history $\mathcal{H}_t = {(x_\tau,y_\tau)}_{\tau=1}^{t}$, can be obtained as $GP_D(\mu_t,k_t(x,x))$ with
\begin{subequations}
\begin{align}
    \mu_t(x)  &=  k_t(x)^{\top} (K_t + \lambda I )^{-1} y_{1:t}, \\
   	k_t(x,x') &= k(x,x') - k_t(x)^{\top} (K_t + \lambda I )^{-1} k_t(x'), \\ 
   	\sigma_t^2(x) &= k_t(x,x),
\end{align}
\label{eq:gaussian_process}
\end{subequations}
where $k_t(x) = [k(x_1,x) \dots k(x_t,x)]^T$.

The algorithm GP bandit algorithms combine the posterior mean $\mu_{t-1}(x)$ and standard deviation $\sigma_{t-1}(x)$ to construct an upper confidence bound (UCB) on the function and choose the action that maximizes this UCB. The action taken can be given by
\begin{align}
    x_t =\underset{x \in \mathcal{D}}{\argmax} (\mu_{t-1}(x) + \beta_t \sigma_{t-1}(x)).
    \label{eq:maximizing_single}
\end{align}
This strikes a natural balance between exploration, where uncertainty is high, and exploitation, where the mean is high. Different algorithms, such as GP-UCB \cite{srinivas2009gaussian} and IGP-UCB \cite{chowdhury2017kernelized}, arise from varying values of the trade-off parameter $\beta_t$, resulting in different regret guarantees. The key approach to bounding the regret involves leveraging the concept of maximum information gain. The rate at which the unknown function can be learned is assessed using a metric known as information gain, which is a fundamental concept in Bayesian experimental design \cite{Chaloner1995}. It quantifies the informativeness of a set of sampling points $\A_{1:t}$ by measuring the mutual information between the function $f$ and the observations $y_{1:t}$. The information gain is defined as
\begin{align}
    I(y_{1:t}; f) = H(y_{1:t})- H(y_{1:t}|f)
\end{align}
where $H$ is the entropy of the random variables. 
This amounts to the reduction in uncertainty about $f$ after making the observations $y_{1:t}$.
The maximum information gain $\gamma_t$ represents the maximum values of information gain for any $\A_{1:t}\in \mathcal{D}$. The maximum information gain only depends on the knowledge of domain $\mathcal{D}$ and the kernel function $k$ and increases only polylogarithmically on time $t$. For squared exponential kernels we have $\gamma_t = \mathcal{O}((\log t)^{d+1})$ \cite{srinivas2009gaussian}, and for Mat\'ern kernels we have $\gamma_t = \mathcal{O}(t^{\frac{d}{2\nu +d}}(\log^\frac{2\nu}{2\nu +d} (t)))$ \cite{vakili2021information}.

\section{Algorithm}
\label{sec:algorithm}

We propose the MA-IGP-UCB algorithm (in Algorithm \ref{alg:1}), which enables each agent to estimate the mean and standard deviation of its local function $f_i$ but explores and exploits the action space based on the estimated confidence bound of the global reward function $F$.
To achieve this, agents need to communicate their estimates of the posterior mean and standard deviation of the global function with their neighbors and reach a consensus on these estimates. However, two main challenges naturally arise in implementing this approach. First, the value for which consensus is required changes over time as agents incrementally learn about their functions. Second, achieving distributed consensus itself necessitates multiple rounds of communication to obtain accurate estimates. To address these challenges, we develop a distributed approach taking inspiration from the running consensus algorithm proposed in \cite{chen2012distributed}.

\begin{algorithm} [t]
\caption{MA-IGP-UCB Algorithm (For agent $i$)}\label{alg:1}
\begin{algorithmic}[1] 
\Require Input space $D$, prior $GP(0,k)$, parameters $B, R, \delta, \lambda$
\For{$t=1,2,3,\dots, T$}
    \State Play $x_{i,t} \leftarrow \underset{x \in D}{\mathrm{argmax}} \; \overline \mu_{i,t-1}(x) + \beta_t \overline \sigma_{i,t-1}(x)$
    \State Observe reward $y_{i,t} = f_i(x_{i,t}) + \epsilon_{i,t}$
    \State Perform Bayesian update to get $\mu_{i,t}$ and $ \sigma_{i,t}$
    \State $\overline \mu_{i,t} \leftarrow (\mu_{i,t}- \mu_{i,t-1})+ \overline \mu_{i,t-1} $\\ \hspace{1.5cm} $+ \sum_{j\in \mathcal{N}_i} w_{ij}(\overline  \mu_{j,t-1} -\overline  \mu_{i,t-1}) $
    \State $\overline \sigma_{i,t} \leftarrow (\sigma_{i,t}- \sigma_{i,t-1})+ \overline \sigma_{i,t-1} $\\ \hspace{1.5cm} $+ \sum_{j\in \mathcal{N}_i} w_{ij}(\overline  \sigma_{j,t-1} -\overline  \sigma_{i,t-1}) $
\EndFor
\end{algorithmic}
\end{algorithm}

Let $A = [w_{ij}] \in \R^{N\times N}$ denote the adjacency matrix associated with the graph $\mathcal{G}$. Metropolis weights are used to define $w_{ij} = \max\{\sfrac{1}{(d_i+1)}, \sfrac{1}{(d_j+1)} \}$ if $(i,j) \in \mathcal{E}$, and $0$ otherwise, where $d_i=|\mathcal{N}_i|$. Based on the actions taken and rewards received, agent $i$ obtains a posterior distribution $\{\mu_{i,t}(x),\sigma_{i,t}(x)\}$ of the local function $f_i$ associated with it using \eqref{eq:gaussian_process}. Given that the objective is to maximize the global function, we define $\overline \mu_{i,t-1}(x)$ and $\overline \sigma_{i,t-1}(x)$ representing the agent $i$'s estimate of the mean and standard deviation of the global function at time step $t-1$. Then agent $i$ updates these estimates using:
\begin{subequations}
\begin{align}
    \overline \mu_{i,t}(x) =& (\mu_{i,t}(x)- \mu_{i,t-1}(x))+ \overline \mu_{i,t-1}(x) \nonumber \\
    &+ \sum_{j\in \mathcal{N}_i} w_{ij}(\overline  \mu_{j,t-1}(x) -\overline  \mu_{i,t-1}(x)), 
    \label{eq:mean_update} \\
    \overline \sigma_{i,t}(x) =& (\sigma_{i,t}(x)- \sigma_{i,t-1}(x))+ \overline \sigma_{i,t-1}(x) \nonumber \\ &+ \sum_{j\in \mathcal{N}_i} w_{ij}(\overline  \sigma_{j,t-1}(x) -\overline  \sigma_{i,t-1}(x)).
    \label{eq:sigma_update} 
\end{align}
\label{eq:all_update}
\end{subequations}
\begin{remark}
    The update laws \eqref{eq:mean_update} and \eqref{eq:sigma_update} capture the spirit of running consensus. The agents want to learn the global function which is the average of local functions. The accuracy of the global function's estimate hinges on two factors: accuracy of the  local function estimates and the accessibility of the other agent's local functions. It's important to note that both these factors vary over time. The first term of \eqref{eq:mean_update} and \eqref{eq:sigma_update} in the parentheses encapsulates the effect of the former by adding the difference between the current and previous estimate. The remaining component captures the influence of the latter through the application of the distributed consensus approach.
\end{remark}

The action for the next time step is determined using \eqref{eq:maximizing_single}, but its evaluation is based on the estimates $\{\overline \mu_{i,t-1}(x),\overline \sigma_{i,t-1}(x)\}$, rather than the estimates of its own local function. Note that the proposed algorithm does not require agents to share information about their actions, rewards, or local function estimates, thereby providing privacy.

\section{Regret Bounds}
\label{sec:regret_bounds}

We now present the regret bound of our MA-IGP-UCB algorithm.

\begin{theorem}
Let $D \subset \R^d$ be a compact and convex action space,  $\delta \in (0,1)$. Under the assumptions (1) and (2), for a connected network of $N$ agents, the MA-IGP-UCB (Algorithm \ref{alg:1}) with $\beta_t  = B + \eta \sqrt{2(\gamma_{t-1}+1+\ln(N/\delta))}$ we obtain a regret bound of $\oo^{*}(N^2\sqrt{T}(B\sqrt{\gamma_T} + \gamma_T))$ with high probability. Precisely, with a probability of at least $1-\delta$, we get
\begin{align}
    R(T) \leq & 2\beta_T \left(1+ \frac{2(N-1)N}{1-|\lambda_2|}\right) \sqrt{4T\lambda\gamma_T} \nonumber \\
    &+  \frac{N(N-1)B|\lambda_2|}{1-|\lambda_2|} + 4B
    \label{eq:thm1_regret}
\end{align}
\label{thm1}
where $\lambda_2$ is the second largest eigenvalue (in absolute value) of the Perron matrix\footnote{Perron matrix is defined as $P=I-L$, where $L$ is the Laplacian of graph $\mathcal{G}$. } of graph $\mathcal{G}$.
\end{theorem}

\begin{remark}
    We note that the obtained regret bounds are sub-linear only for those where $\gamma_T$ grows faster than $\sqrt{T}$. This includes squared exponential kernel and Matérn kernel with $\nu>d/2$.
\end{remark}

Our proposed algorithm, utilizes the estimates of global function to construct confidence intervals, building up on the IGP-UCB algorithm \cite{chowdhury2017kernelized}. The major challenge in proving a sub-linear regret bound is caused by the heterogeneity of actions among the agents. To understand this, we observe that when calculating the instantaneous regret, we sum the equation \eqref{eq:mean_update} and \eqref{eq:sigma_update} for all the agents, which are evaluated at different values $\{x_{i,t}\}$. This introduces a Hadamard product between the Perron matrices and the matrices representing the estimated mean and standard deviation of all local functions evaluated at all values of $\{x_{i,t}\}$. This novel aspect of our analysis is further explained in the supplementary material, which provides a comprehensive proof.
The per-agent regret bound derived exhibits a dependence on $N^2$ in the second term. The heterogeneity of actions as discussed above contributes to the first $N$. The second $N$ arises when bounding the convergence of the Perron matrix. As the regret bound depends on $\lambda_2$, a natural question arises: what is the best the algorithm can do when the network is completely connected?

\begin{corollary}
[Completely connected graph] Under the conditions of Theorem \ref{thm1}, if the communication network of agents is completely connected, then MA-IGP-UCB (Algorithm \ref{alg:1})  with $\beta_t  = B + \eta \sqrt{2(\gamma_{t-1}+1+\ln(N/\delta))}$ we obtain a regret bound of $\oo^{*}(\sqrt{T}(B\sqrt{\gamma_T}+\gamma_T))$ with high probability. Precisely, with a probability of at least $1-\delta$, we get
\begin{align}
    R(T) \leq & 4\beta_T \sqrt{4T\lambda\gamma_T} +4B.  \label{eq:cor_regret}
\end{align} \label{cor1}
\end{corollary}
For this simplest scenario of the MA-IGP-UCB algorithm, the regret bound closely matches the one derived in \cite{chowdhury2017kernelized}\footnote{Since there are N agents, we use the result from \cite{chowdhury2017kernelized} with a probability of at least $1- \delta/N$.}.

\begin{remark}
    Lower bound: One trivial approach to examining the lower limit of cumulative regret for the MA-IGP-UCB algorithm is to consider the best case scenario of a completely connected setting. This aligns with the single-agent setting established in Corollary 1. This allows us to directly apply the findings from \cite{scarlett2017lower}, resulting in $\Omega(T(\log(T))^d)$ for squared exponential kernel and $\Omega(T^{\frac{\nu + d}{2\nu + d}})$ for Matérn kernel. The same optimality outcomes are evident, as presented in \cite{scarlett2017lower}. However, this approach is quite conservative for the distributed scenario as it doesn't encompass the network structure. A tighter lower bound for distributed setting remains a potential avenue for future exploration. 
\end{remark}

\section{Extension}
\label{sec:extension}

In this section, we present an extension of the MA-IGP-UCB algorithm (Algorithm \ref{alg:1}) by drawing inspiration from the work in \cite{martinez2019decentralized}. The problem with the MA-IGP-UCB algorithm is that it takes time for the estimates to get communicated across the network. Also, as seen from \eqref{eq:mean_update} and \eqref{eq:sigma_update}, the most recent updates in $\mu_{i,t}$ and $\sigma_{i,t}$ (the agent's estimate of the local functions) strongly influence their estimate of the global function. The influence is more pronounced if the spectral gap is smaller.  One solution to this issue is to exclude recent estimates when constructing the upper confidence bound. Authors in \cite{martinez2019decentralized} proposed a similar approach in their work on solving the distributed stochastic bandit problem, where they introduced the Decentralized Delayed Upper Confidence Bound (DDUCB) algorithm. They performed running consensus in stages of a certain number of iterations, incorporating only those estimates that have communicated (mixed) sufficiently. Inspired by their work, we propose an extension called MAD-IGP-UCB (Multi-agent Delayed Improved Gaussian Process Upper Confidence Bound). Similar to DDUCB, the consensus in MAD-IGP-UCB is done in stages and numerically outperforms MA-IGP-UCB, but at the cost of doubling communication.

We now detail the MAD-IGP-UCB (Algorithm \ref{alg:2})for the whole multi-agent system. The algorithm operates in stages, each of which consists of $c$ iterations. At any stage $s$, we have three sets of estimates: the \textit{running estimate}, the \textit{mixing estimate}, and the \textit{mixed estimate}. The \textit{running estimate} of the mean and standard deviation of the global function for agent $i$ is the same in MA-IGP-UCB denoted by $\{\overline \mu_{i,t}, \overline \sigma_{i,t}\}$. If the current time step $t$ is in stage $s$, let $t_s = (s-1)c$ denote the last time step of the previous stage. At the beginning of stage $s$, the \textit{mixing estimate} $\{\overline \mu_{mg,i,t}, \overline \sigma_{mg,i,t}\}$ are initialized with the value $\{\overline \mu_{i,t_s}, \overline \sigma_{i,t_s}\}$, and are updated during the stage using the distributed consensus algorithm. The \textit{mixed estimates} $\{\overline \mu_{md,i,t}, \overline \sigma_{md,i,t}\}$ denote the estimates that have been mixed at least $c$ times and correspond the actual estimates that were calculated until time $t_s-c$. These mixed estimates are used by the agents to decide which action to take.

\begin{algorithm}[t]
\caption{MAD-IGP-UCB Algorithm (For agent $i$)}\label{alg:2}
\begin{algorithmic}[1]
\Require Input space $D$, prior $GP(0,k)$, parameters $B, R, \delta, \lambda, \epsilon, c$,
\For{$s=1,2,3,\dots, \lfloor T/c \rfloor$} \Comment{s denotes stages}
\State $t_s \leftarrow sc$
\If{$s>=3$} 
\State $\overline \mu_{md,i,t_s} \leftarrow \overline \mu_{mg,i,t_s}$
\State $\overline \sigma_{md,i,t_s} \leftarrow \overline \sigma_{mg,i,t_s}$ \Comment{Updating mixed est.}
\EndIf
\If{$s>=2$}
\State $\overline \mu_{mg,i,t_s} \leftarrow \overline \mu_{i,t_s}$
\State $\overline \sigma_{mg,i,t_s} \leftarrow  \overline \sigma_{i,t_s}$ \Comment{Updating est. to be mixed}
\EndIf
\State $\tau \leftarrow 0$ 
\While{$\tau < c$}
    \State $t \leftarrow t_s + \tau$
    \State Play $x_{i,t} \leftarrow \underset{x \in D}{\mathrm{argmax}} \; \overline \mu_{md,i,t_s}(x) + \beta_{t_s} \overline \sigma_{md,i,t_s}(x)$
    \State Observe reward $y_{i,t} = f_i(x_{i,t}) + \epsilon_{i,t}$
    \State Perform Bayesian update to get $\mu_{i,t}$ and $ \sigma_{i,t}$
    \State $\overline \mu_{i,t} \leftarrow (\mu_{i,t}- \mu_{i,t-1})+ \overline \mu_{i,t-1} $\\ \hspace{2cm} $+ \sum_{j\in \mathcal{N}_i} w_{ij}(\overline  \mu_{j,t-1} -\overline  \mu_{i,t-1}) $ 
    \State $\overline \sigma_{i,t} \leftarrow (\sigma_{i,t}- \sigma_{i,t-1})+ \overline \sigma_{i,t-1} $\\ \hspace{2cm} $+ \sum_{j\in \mathcal{N}_i} w_{ij}(\overline  \sigma_{j,t-1} -\overline  \sigma_{i,t-1}) $
    \State $\overline \mu_{mg,i,t} \leftarrow \overline \mu_{mg,i,t-1} $\\ \hspace{2cm} $+ \sum_{j\in \mathcal{N}_i} w_{ij}(\overline  \mu_{mg,j,t-1} -\overline  \mu_{mg,i,t-1}) $ 
    \State $\overline \sigma_{mg,i,t} \leftarrow \overline \sigma_{mg,j,t-1} $\\ \hspace{2cm} $+ \sum_{j\in \mathcal{N}_i} w_{ij}(\overline  \sigma_{mg,j,t-1} -\overline  \sigma_{mg,i,t-1}) $
    \State $\tau \leftarrow \tau+1$
\EndWhile
\EndFor
\end{algorithmic}
\end{algorithm}

\begin{theorem}
Let $D \subset \R^d$ be a compact and convex action space,  $\delta \in (0,1)$. Under the assumptions (1) and (2), for a connected network of $N$ agents, the MAD-IGP-UCB (Algorithm \ref{alg:2}) with $\beta_t  = B + \eta \sqrt{2(\gamma_{t-1}+1+\ln(N/\delta))}$ we obtain a regret bound of $\oo^* \lb \lb N+N^2|\lambda_2|^{c}\rb\sqrt{T}(B\sqrt{\gamma_T}+\gamma_T) \rb$ with high probability. More precisely. with a probability of at least $1-\delta$, we get
\begin{align}
R(T) \leq &  2\beta_T \left(N+ \frac{2(N-1)N|\lambda_2|^{c}}{1-|\lambda_2|}\right) \sqrt{4T\lambda\gamma_T} \nonumber \\
    &+  \frac{N(N-1)B|\lambda_2|^{2c}}{1-|\lambda_2|} + 4cB
\label{eq:thm2_regret}
\end{align} 
where $\lambda_2$ is the second largest eigenvalue (in absolute value) of the Perron matrix of graph $\mathcal{G}$.
\label{thm2}
\end{theorem}

By comparing the per-agent regret bound obtained in Equation \eqref{eq:thm1_regret}, we can observe that MAD-IGP-UCB significantly reduces the dependence on the number of agents through the term $|\lambda_2|^c$. This reduction is attributed to the fact that the estimates are communicated and mixed at least $c$ times before being used for decision-making. The algorithm introduces an extra $N$ in the first term due to the heterogeneity of the action. We note that the algorithm is sensitive to the choice of $c$. Using a large value of $c$ can lead to two main issues. Firstly, the agents may use estimates with a delay of $2c-1$ steps, resulting in incomplete utilization of new information. Secondly, the agents' actions remain fixed during a stage, resulting in constant regret. A well-balanced value of $c$ would provide a more accurate estimate of the global function with sufficient mixing and minimal delay.

\begin{figure}[t] 
    \centering \includegraphics[width=0.6\linewidth, trim={1.1cm 0.7cm 1cm 0.5cm}, clip]{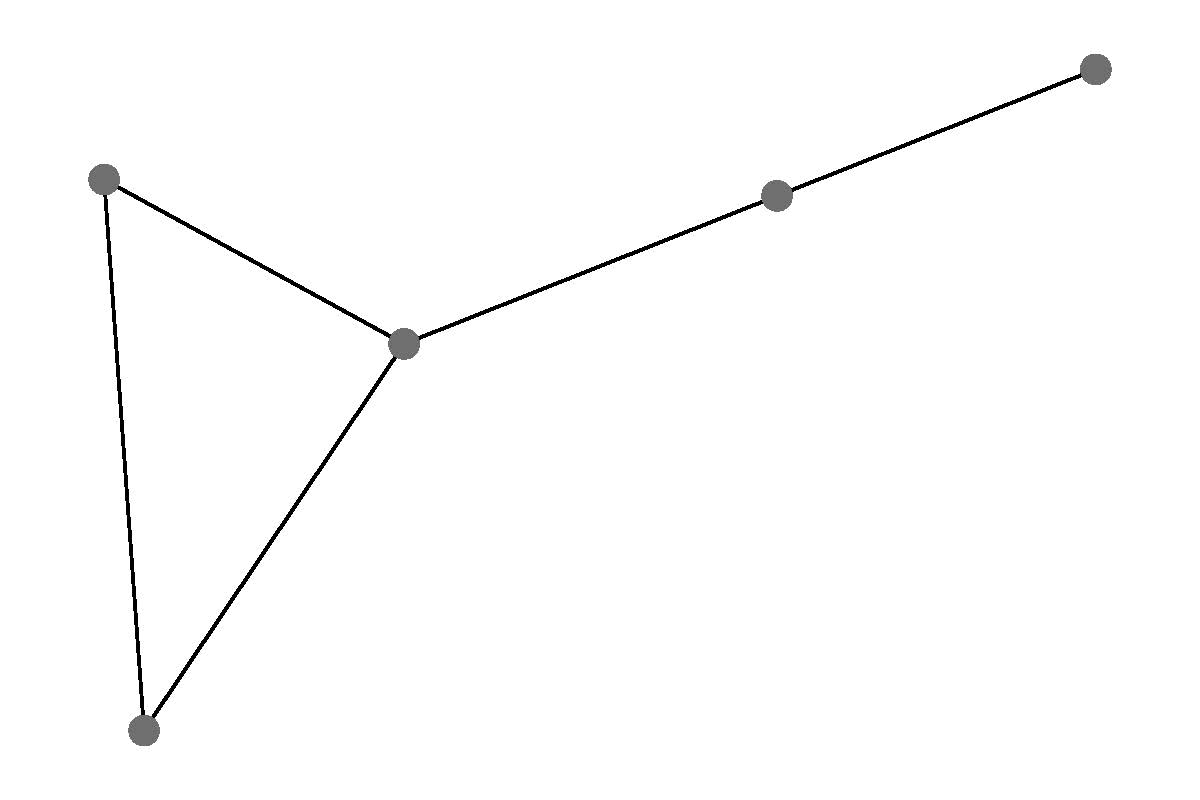}
        \caption{Graph of 5 agent network}
        \label{fig:net_conn}
\end{figure}

\begin{figure*}[t!] 
    \centering
    \begin{subfigure}{0.328\textwidth}
        \centering
        \includegraphics[width=\linewidth, trim={0.44cm 0.35cm 1.6cm 1.4cm}, clip]{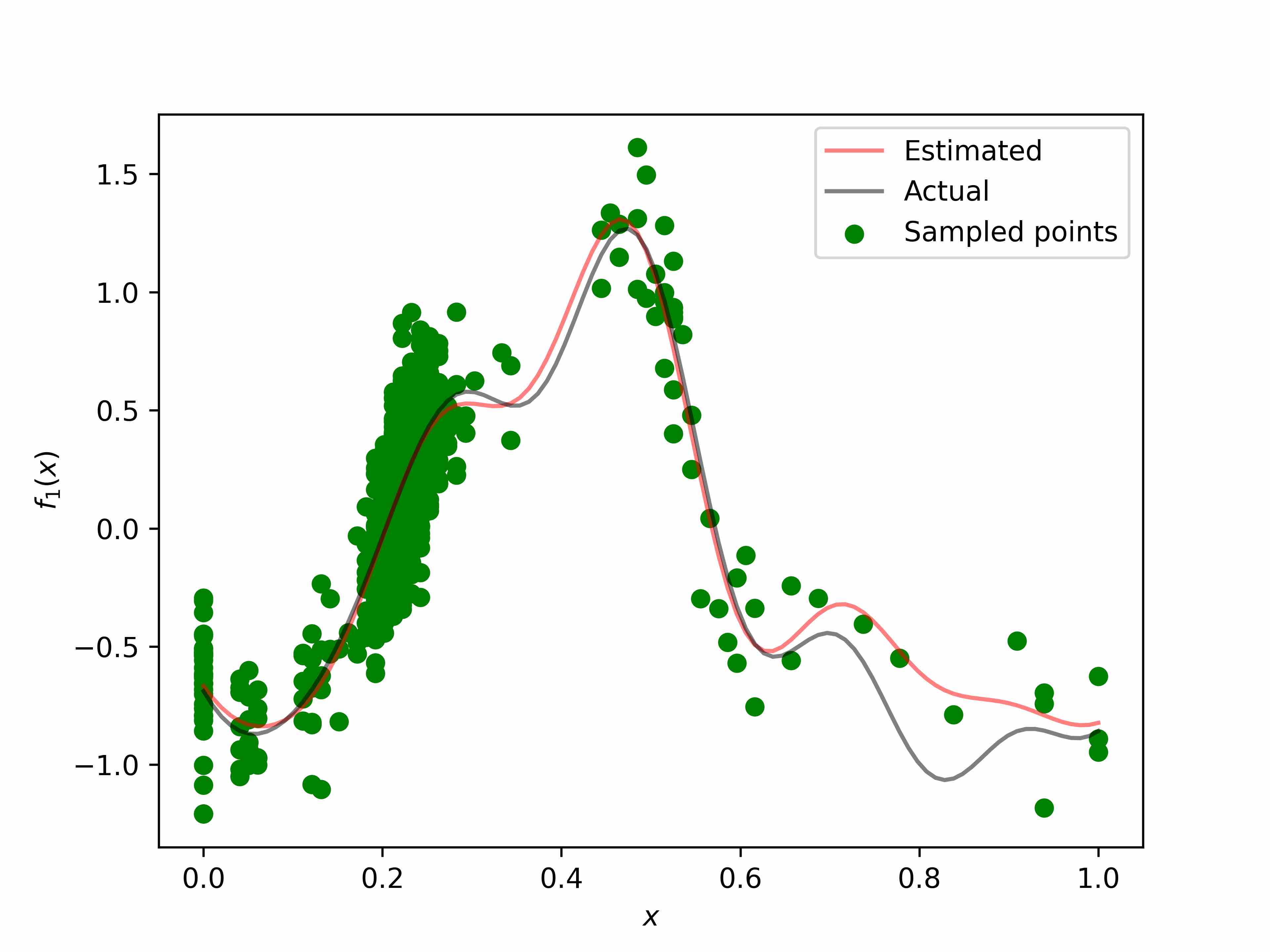}
        \caption{Agent $1$ local function}
        \label{fig:pri_a1}
    \end{subfigure}
    \hfill
    \begin{subfigure}{0.328\textwidth}
        \centering
        \includegraphics[width=\linewidth, trim={0.44cm 0.35cm 1.6cm 1.4cm}, clip]{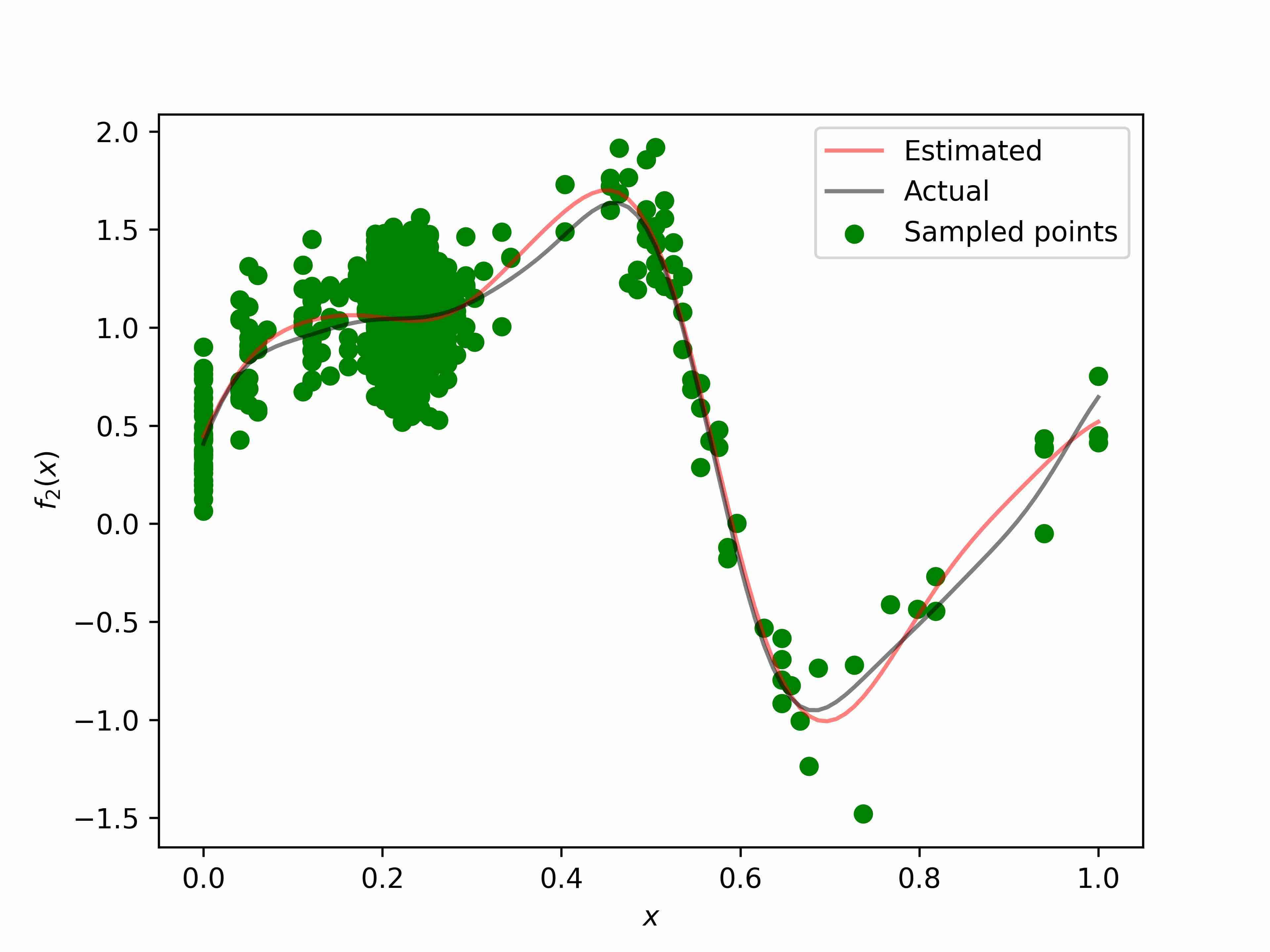}
        \caption{Agent $2$ local function}
        \label{fig:pri_a2}
    \end{subfigure}
    \hfill
    \begin{subfigure}{0.328\textwidth}
        \centering
        \includegraphics[width=\linewidth, trim={0.44cm 0.35cm 1.6cm 1.4cm}, clip]{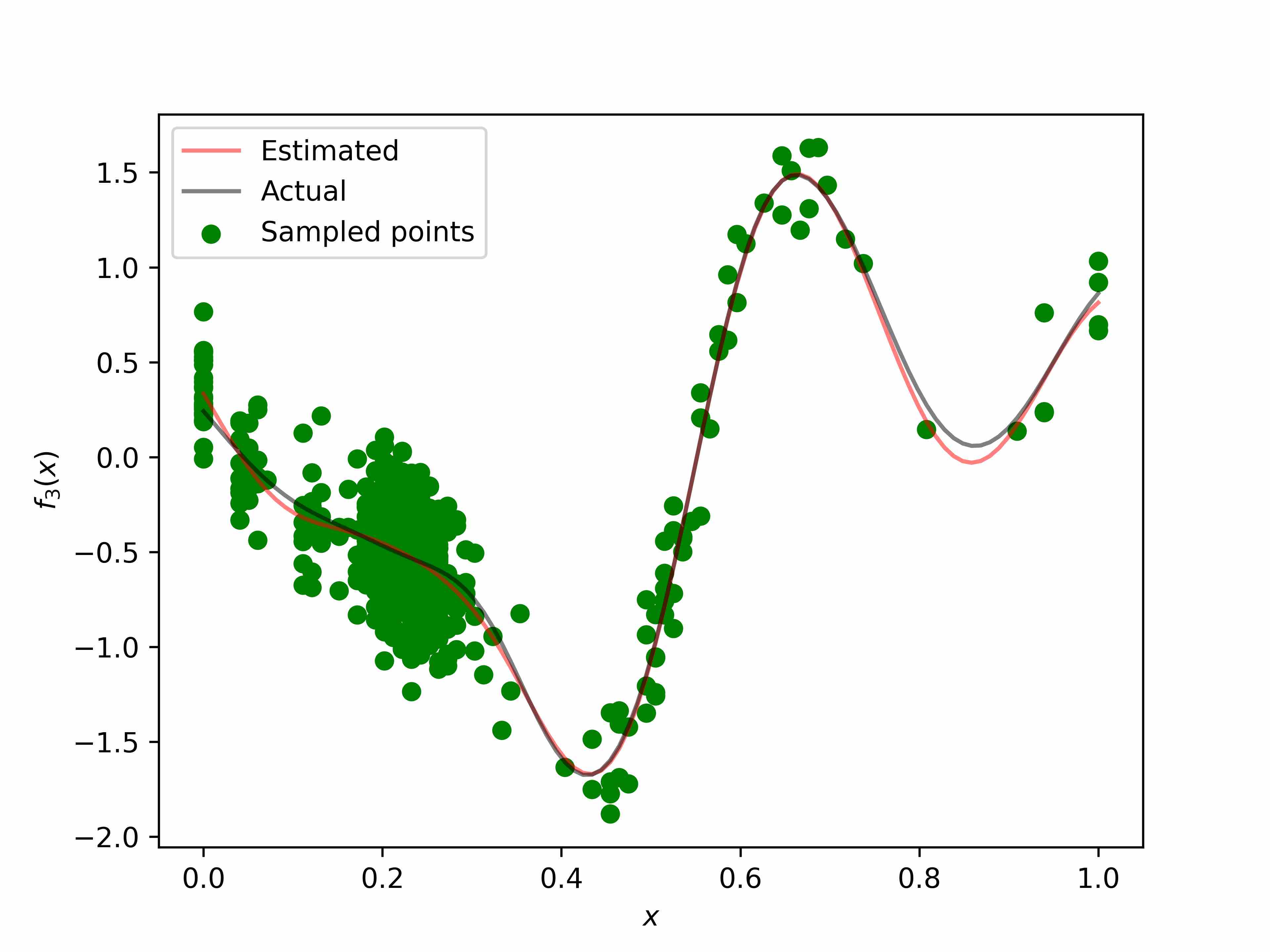}
        \caption{Agent $3$ local function}
        \label{fig:pri_a3}
    \end{subfigure}

    \begin{subfigure}[b]{0.328\textwidth}
        \centering
        \includegraphics[width=\linewidth, trim={0.44cm 0.35cm 1.6cm 1.4cm}, clip]{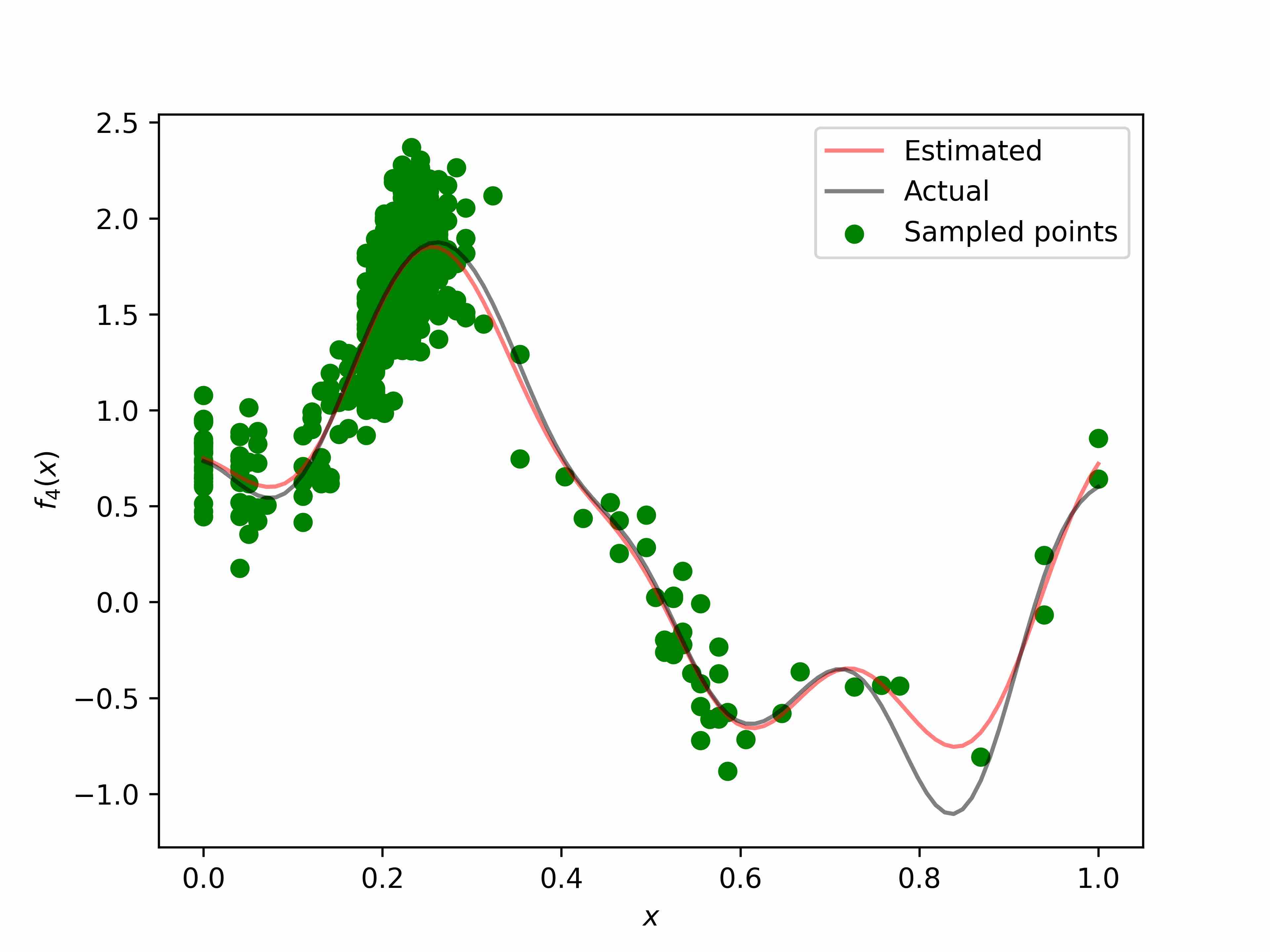}
        \caption{Agent $4$ local function}
        \label{fig:pri_a4}
    \end{subfigure}
    \hfill
    \begin{subfigure}[b]{0.328\textwidth}
        \centering
        \includegraphics[width=\linewidth, trim={0.44cm 0.35cm 1.6cm 1.4cm}, clip]{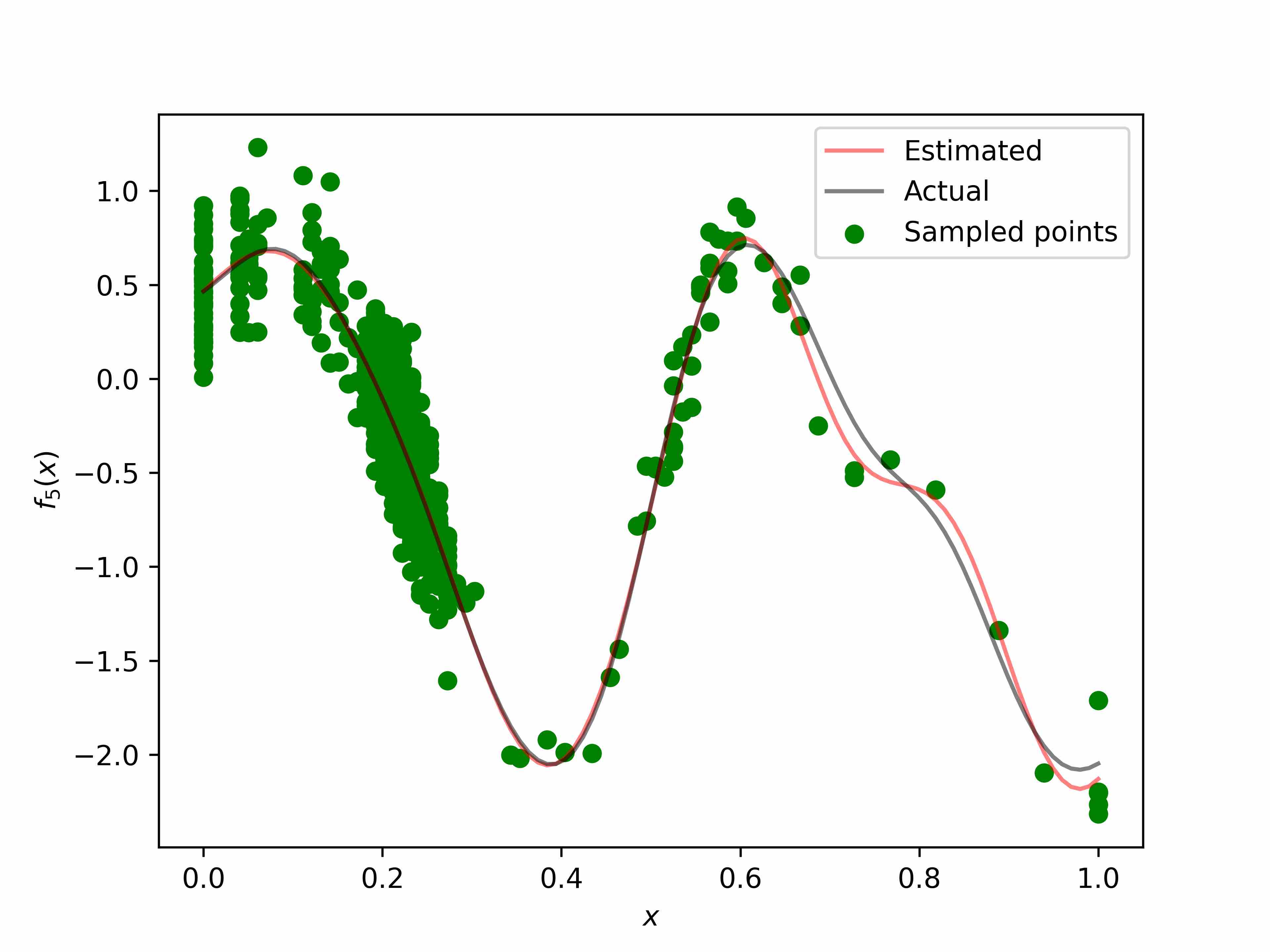}
        \caption{Agent $5$ local function}
        \label{fig:pri_a5}
    \end{subfigure}
    \hfill
    \begin{subfigure}[b]{0.328\textwidth}
        \centering
        \includegraphics[width=\linewidth, trim={0.44cm 0.35cm 1.6cm 1.4cm}, clip]{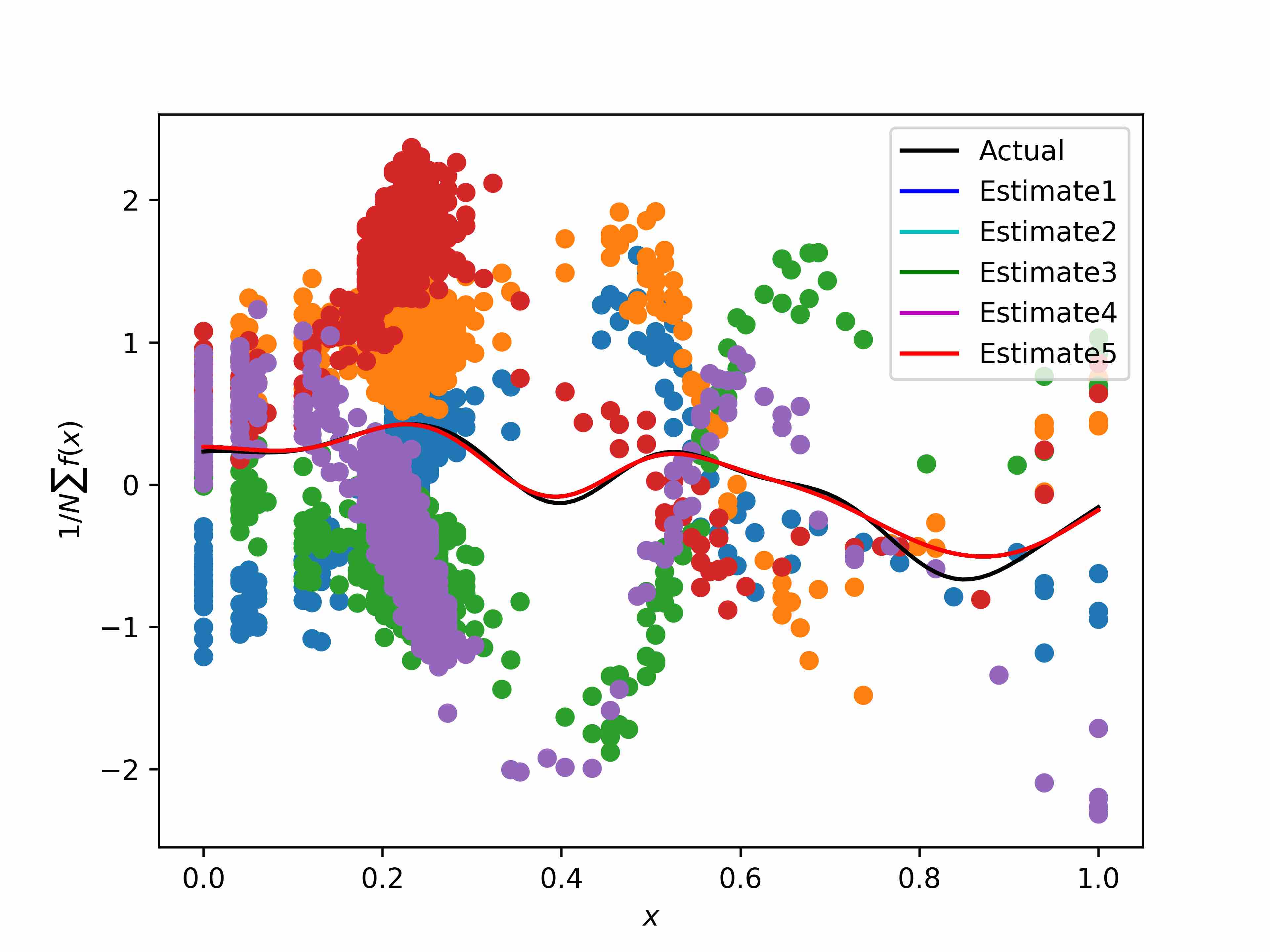}
        \caption{Global function}
        \label{fig:pri_global}
    \end{subfigure}
    \caption{Comparison of actual functions versus posterior mean estimates with sampling points (T = $1000$ iterations) using the MA-IGP-UCB algorithm. Subfigures (a)-(e) depict the local functions, while subfigure (f) presents the actual global function along with each agent's estimate of the global function.}
    \label{fig:pri_functions}
\end{figure*}

\begin{figure*}[htbp!] 
    \centering
    \begin{subfigure}{0.328\textwidth}
        \centering
        \includegraphics[width=\linewidth, trim={0.44cm 0.35cm 1.6cm 1.4cm}, clip]{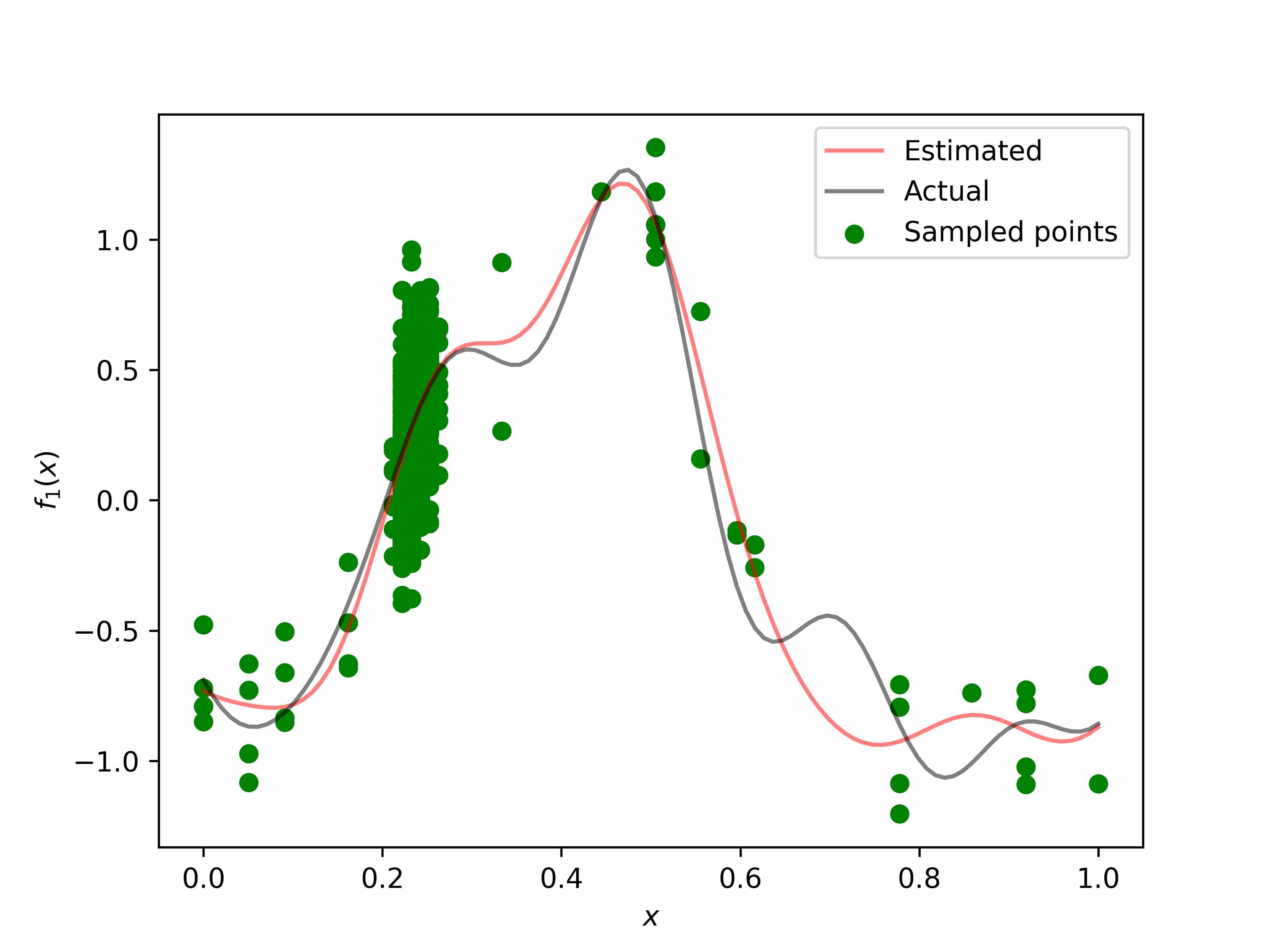}
        \caption{Agent $1$ local function}
        \label{fig:del_a1}
    \end{subfigure}
    \hfill
    \begin{subfigure}{0.328\textwidth}
        \centering
        \includegraphics[width=\linewidth, trim={0.44cm 0.35cm 1.6cm 1.4cm}, clip]{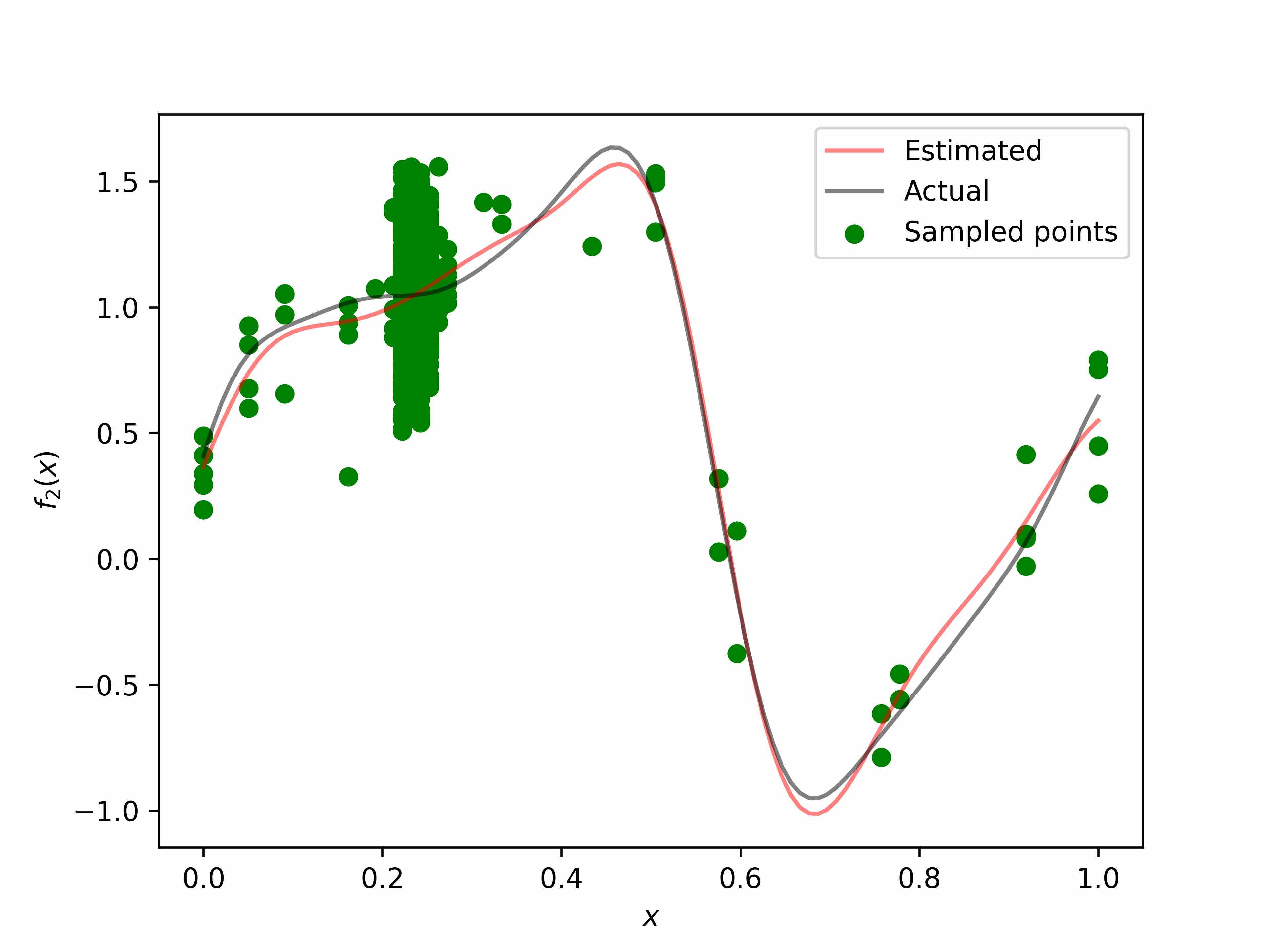}
        \caption{Agent $2$ local function}
        \label{fig:del_a2}
    \end{subfigure}
    \hfill
    \begin{subfigure}{0.328\textwidth}
        \centering
        \includegraphics[width=\linewidth, trim={0.44cm 0.35cm 1.6cm 1.4cm}, clip]{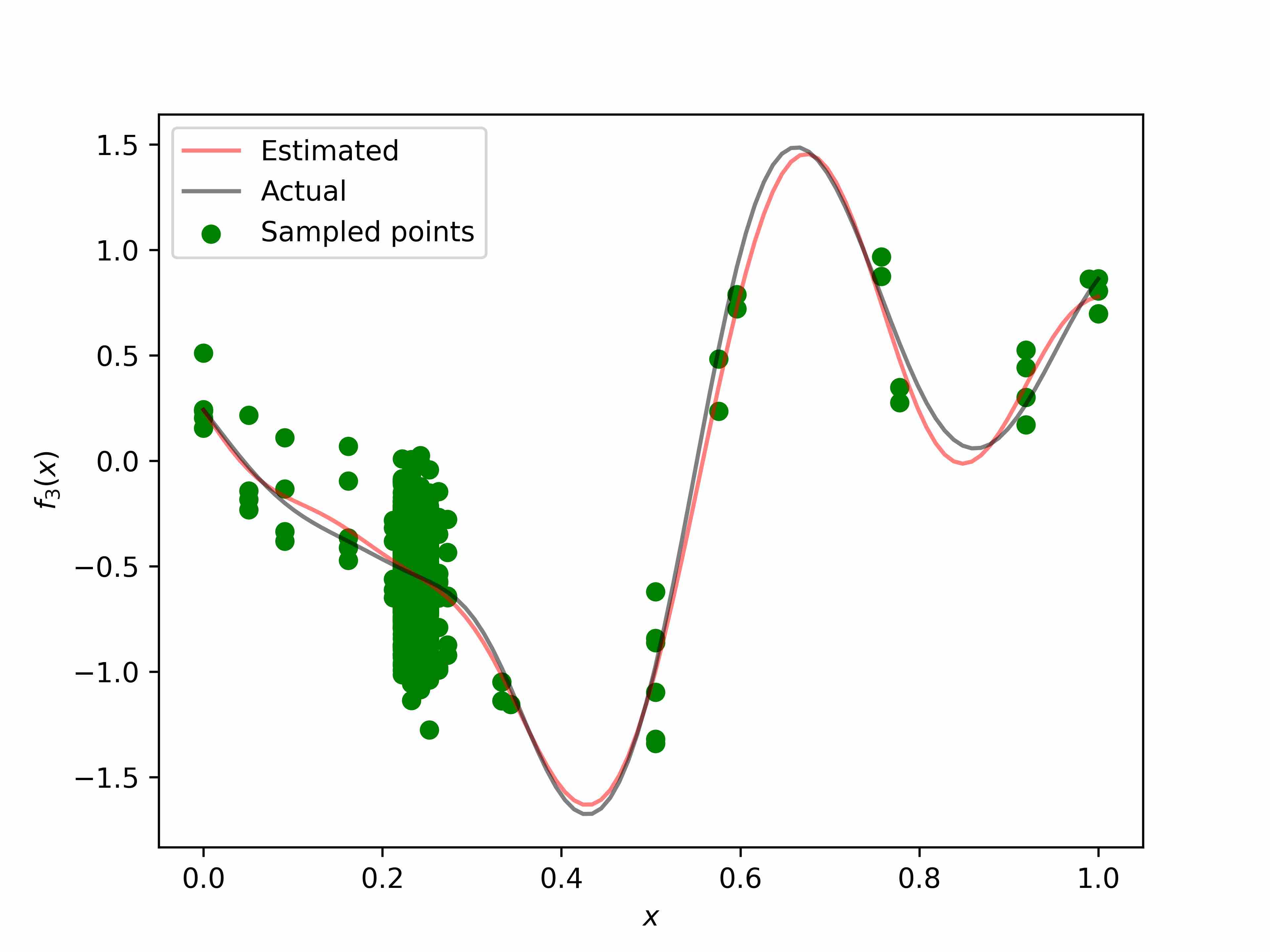}
        \caption{Agent $3$ local function}
        \label{fig:del_a3}
    \end{subfigure}

    \begin{subfigure}{0.328\textwidth}
        \centering
        \includegraphics[width=\linewidth, trim={0.44cm 0.35cm 1.6cm 1.4cm}, clip]{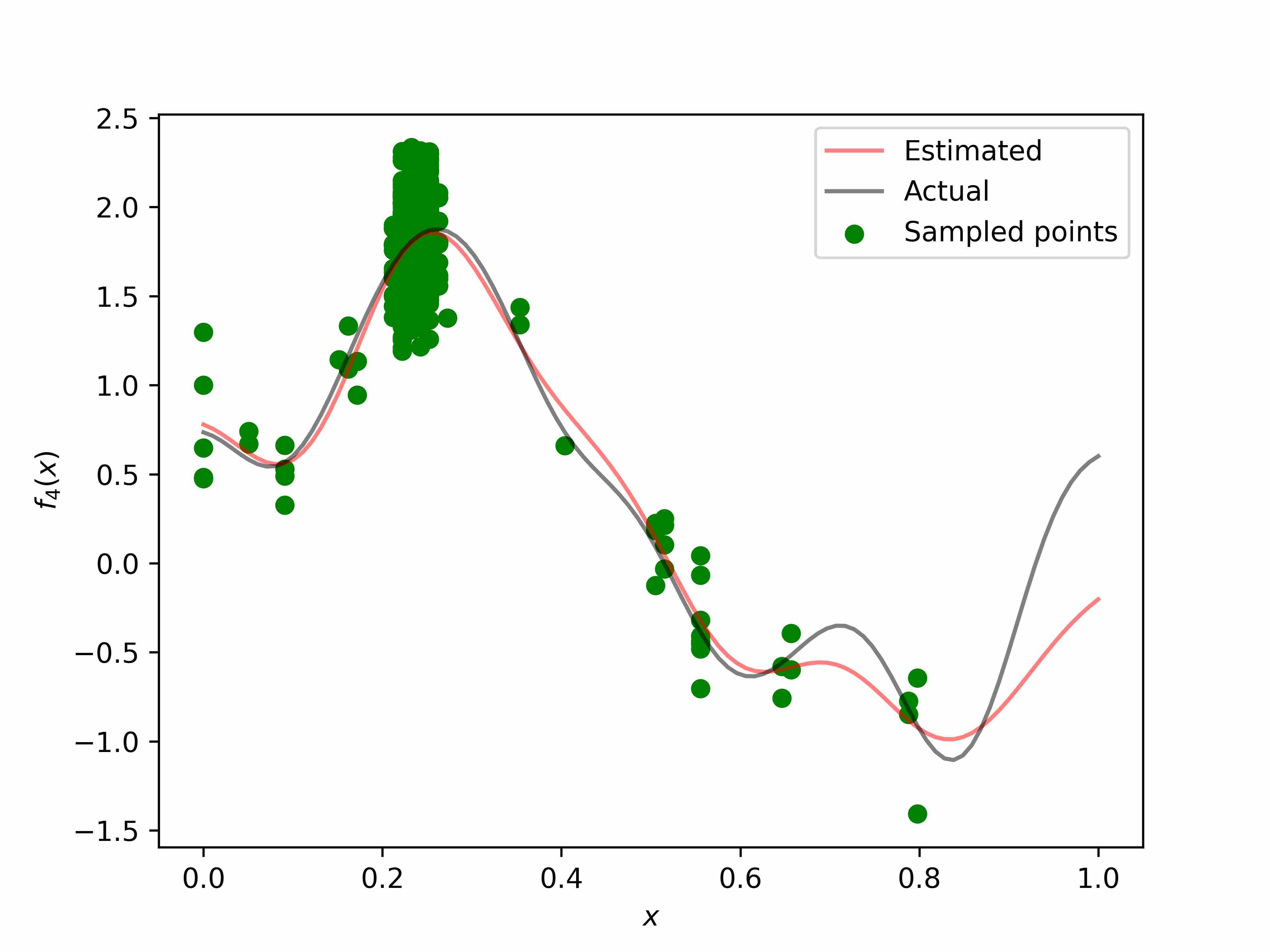}
        \caption{Agent $4$ local function}
        \label{fig:del_a4}
    \end{subfigure}
    \hfill
    \begin{subfigure}{0.328\textwidth}
        \centering
        \includegraphics[width=\linewidth, trim={0.44cm 0.35cm 1.6cm 1.4cm}, clip]{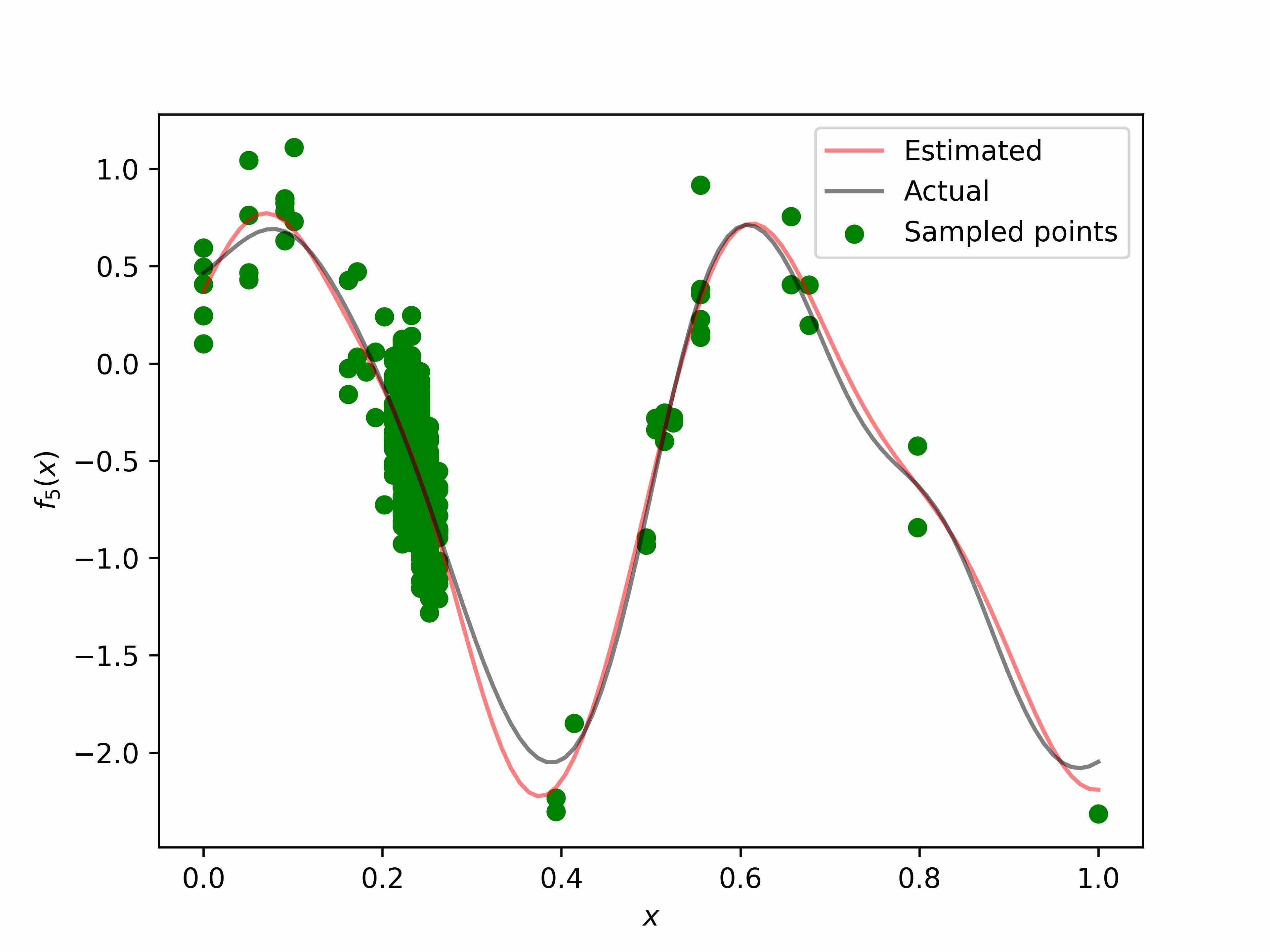}
        \caption{Agent $5$ local function}
        \label{fig:del_a5}
    \end{subfigure}
    \hfill
    \begin{subfigure}{0.328\textwidth}
        \centering
        \includegraphics[width=\linewidth, trim={0.44cm 0.35cm 1.6cm 1.4cm}, clip]{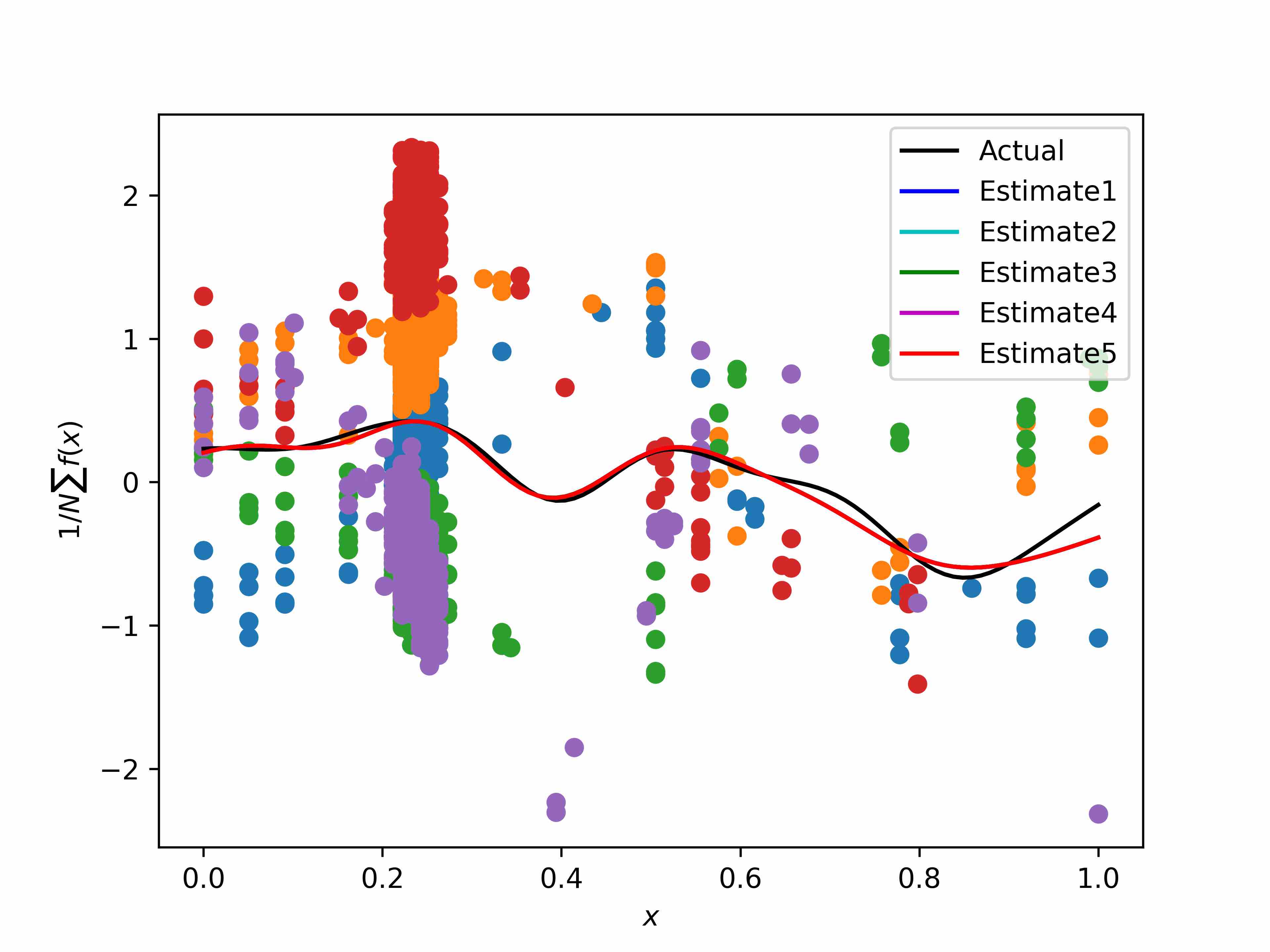}
        \caption{Global function}
        \label{fig:del_global}
    \end{subfigure}
    \caption{Comparison of actual functions versus posterior mean Estimates with sampling points using the MAD-IGP-UCB algorithm.  Subfigures (a)-(e) depict the local functions, while subfigure (f) presents the actual global function along with each agent's estimate of the global function.}
    \label{fig:del_functions}
\end{figure*}

\begin{figure*}[t!]
    \centering
    \includegraphics[width=0.95\linewidth,trim={2.5cm 1cm 3.25cm 0.cm},clip]{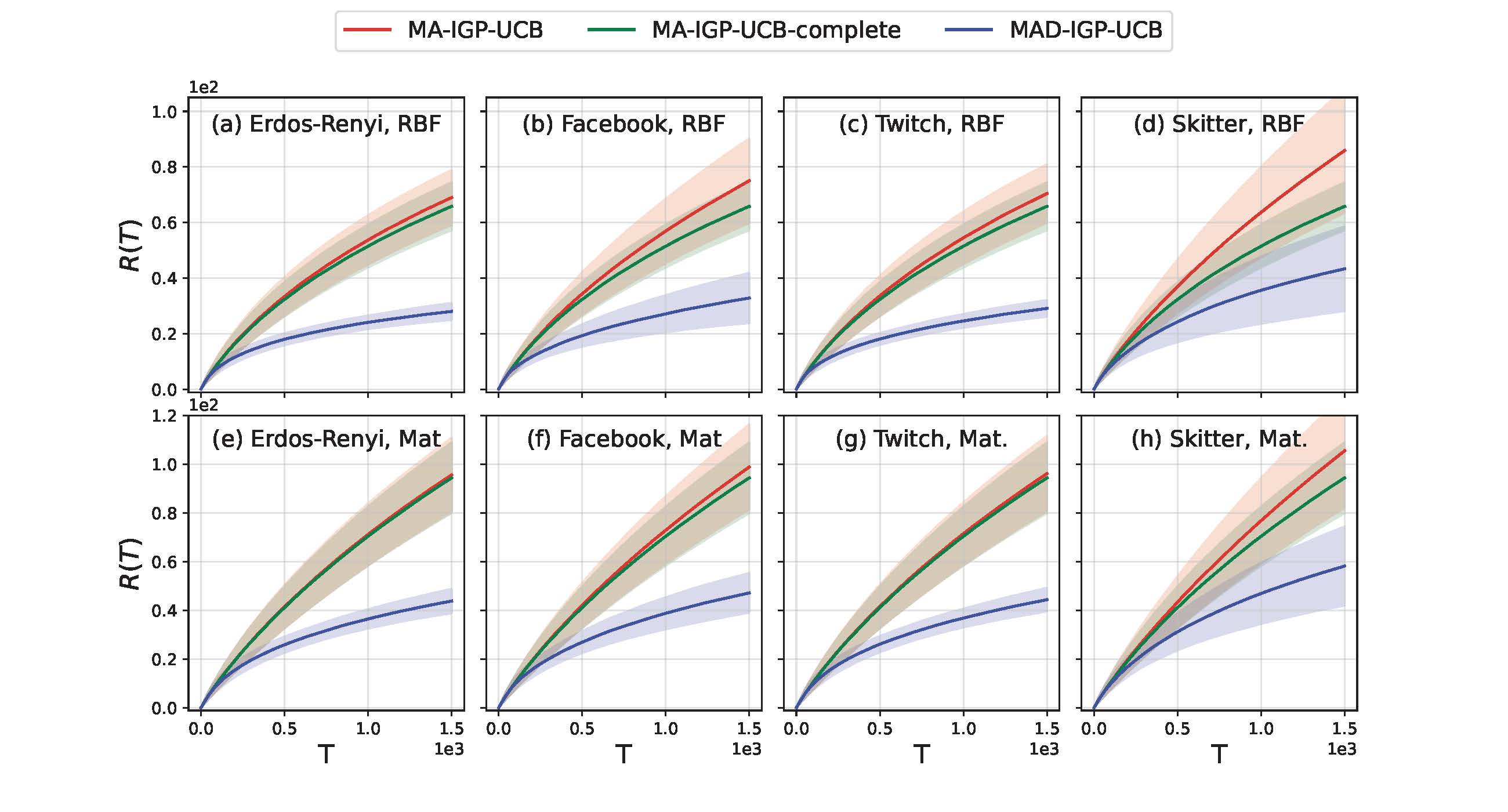}
    \caption{Comparison of numerical results for the distributed optimization algorithms: MA-IGP-UCB, MA-IGP-UCB in completely connected setting, and MAD-IGP-UCB. The results are averaged over $100$ independent trials, where the mean is displayed in bold, and the standard deviation is indicated using the shaded region.}
    \label{plot}
\end{figure*}

\section{Experiments}
\label{sec:experiments}

This section presents the numerical results of the proposed algorithms using synthetic test functions on small networks, randomly generated large networks as well as networks sub-sampled from real-world networks. The action space $\mathcal{D}$ is obtained by selecting $100$ points uniformly from an interval $[0,1]$. For each agent, the test function is generated by sampling a multi-variate Gaussian distribution with mean $0$ and kernel matrix $K$ calculated on those points. The noise parameters are set to $\eta = 0.2$ and $\lambda = 0.04$. We use Squared exponential and Mat\'ern kernels ($\nu$ = 2.5), both with a lengthscale parameter of $l=0.1$, resulting in highly non-convex functions. For the algorithms MA-IGP-UCB and  MAD-IGP-UCB, we calculate $\gamma_t$ using the theoretical upper bound. The $\beta_t$ is scaled down compared to theoretical values to decrease the convergence time and $c=2$ is used. 

\textit{a) Small Networks: } To study the performance and sampling pattern, we first analyse the experimental results for a small network of $5$ agents (Fig. \ref{fig:net_conn}). We study and compare each agent's estimates of the local functions, estimates of the global function, and the sampled points. The results for the squared exponential kernel using the MA-IGP-UCB algorithm are depicted in Fig. \ref{fig:pri_functions}, while the results for the MAD-IGP-UCB algorithm are shown in Fig. \ref{fig:del_functions} where we plot the actual functions, estimated functions, and noisy sampled points. Both algorithms demonstrate the agents' ability to learn and select actions that maximize the global function. As depicted in Fig. \ref{fig:pri_global} and Fig. \ref{fig:del_global}, the agents reach a consensus on their estimates of the global function, which closely aligns with the actual global function at the maximizing point. As anticipated, the MAD-IGP-UCB algorithm yields superior outcomes owing to the delayed but improved estimates.

\textit{a) Large Networks: }For large networks, we conduct experiment with $100$ agents is run for $1500$ iterations and averaged over $100$ independent trials. During the initial two stages of MAD-IGP-UCB, the MA-IGP-UCB algorithm is employed instead, as no learning occurs and the running estimates are readily available. This is because the agents already possess running estimates that are readily available to them.

\textbf{Network Structure.} We evaluate the proposed algorithms on two types of network structures: synthetic and real-world networks.\footnote{For implementing the MA-GP-UCB algorithm in a completely connected setting, we assume complete connection for all the networks.} For synthetic networks, we generate a connected Erdos-Renyi network of size $N=100$ with an edge probability of $p=0.04$ (sparsely connected). For real-world networks, we randomly sample a connected subgraph from the SNAP network datasets \cite{leskovec2016snap}, which include the ego-Facebook, musae-Twitch, and as-Skitter networks that are common benchmarks for distributed algorithms \cite{dubey2020kernel,dubey2020cooperative}.

\textbf{Results.}
The results of our experiments are presented in Fig. \ref{plot}, where all the algorithms exhibit sub-linear growth of cumulative regret, consistent with our theoretical results. Across all networks and kernels, we find that MAD-IGP-UCB outperforms MA-IGP-UCB, even in the centralized setting. This is likely because MA-IGP-UCB takes a biased decision due to the most recent update of its local estimate, which is prevented in MAD-IGP-UCB due to the mixed estimate. It is worth noting that for the completely connected network, the performance of the MA-IGP-UCB algorithm closely resembles that in the sparsely connected Erdos-Renyi network with an edge probability of $p=0.04$ (see Fig. \ref{plot}a, \ref{plot}e). This similarity can be attributed to the uniform distribution of edges in Erdos-Renyi networks, which leads to a lower value of $\lambda_2$. However, this is not necessarily the case for real-world networks despite having more number of edges.

\section{Discussion and Conclusion}
\label{sec:conclusion}

In this work, we presented a kernelized multi-armed bandit approach for solving distributed optimization problems with non-convex, independent, and unknown functions with low-norm in the RKHS. We proposed two decentralized algorithms, MA-IGP-UCB and MAD-IGP-UCB. These algorithms have been designed to operate in a fully distributed manner, with agents only required to share their estimates about the global function with their immediate neighbors. This property promotes privacy to some extent and makes these algorithms suitable for various practical applications. These algorithms leverage the concept of running consensus to estimate the upper confidence bound on the global function for each agent, which is used to decide the action to be taken.

In contrast to the centralized kernelized bandit problem or distributed optimization problem with known and convex functions, the problem at hand introduces unique challenges. Also, our approach of combining running consensus and Gaussian processes, introduces distinctive complexities in the proof process. We compare the technical challenges on three ends: kernalized bandit approach, running consensus, and delayed estimates. (1) Compared to single-agent case, an estimate of the global function is used to make the decision instead of local on. The analysis warrants to capture the evolution of the global function estimate in terms a local function estimates (from which sampling is done), and further how the standard deviation can be bounded which are evaluated at different values of input. (2) When comparing to the running consensus studied in \cite{chen2012distributed}, their proof relies on the fact that the sum of all estimates is always the the sum of all reference signals. In our regret analysis, this assumption doesn't apply since each agent's action is influenced by its own estimate, leading to the summation of different signals. The tools and techniques used are completely different. (3) Finally, \cite{martinez2019decentralized} which motivates our MAD-IGP-UCB algorithm, introduced a delayed approach but for multi-armed bandits, resulting in a  entirely distinct analysis and proofs. The theoretical outcomes, proof and the integration of distributed consensus within a Gaussian process framework represent the novelty of this paper.

\textbf{Future work.} One limitation of our work is that the theoretical bounds we derived for MAD-IGP-UCB are not sufficiently tight to fully explain its superior numerical performance compared to MA-IGP-UCB. Additionally, the success of our proposed algorithms is highly dependent on the connectivity of the communication graph of the multi-agent system. As an immediate future direction for research, it would be worthwhile to develop algorithms that can handle time-varying graphs, random graphs with weak connectivity, and graphs with asynchronous communication. Security is a critical concern in many applications, and hence, developing differentially private algorithms would be very promising in these settings. In decentralized multi-armed bandit cases, federated approaches have been implemented in \cite{zhu2023distributed, huang2021federated, dubey2020differentially}. Inspired by \cite{bogunovic2016}, we conjecture that an extension to time-varying bandit settings could be developed to solve dynamic optimization problems.

\appendix


\begin{lemma} Under Assumptions 1 and 2, for any $x\in \mathcal{D} \subset \mathbb{R}^n$ and $t\geq1$, it can be guaranteed with a probability of at least $1-\delta$ that $\forall \;\; i=1,\dots,N$
\begin{align}
    |\mu_{i,t}(x) -f_i(x)| \leq \sigma_{i,t}(x)\left(B+\eta \sqrt {2(\gamma_{t}+1+ln(N/\delta))}\right),
    \label{eq_lemma1}
\end{align}
where $\gamma_{t}$ is the maximum information gain after $t$ time steps and $\mu_{i,t}(x)$, $\sigma_{i,t}(x)$ are the mean and standard deviation of the posterior distribution. 
\label{lemma1}
\end{lemma}

\smallskip

\noindent \textbf{Proof of Lemma \ref{lemma1}.}
This result directly extends from Theorem 2, of \cite{chowdhury2017kernelized} by using the probability $\delta/N$ for each agent $i = 1,\dots,N$. Note that $\gamma_{t}$ is the same for all the functions as the maximum information only depends on the action space and kernel function which is the same for all the agents.
\hfill $\qed$

\medskip

\begin{lemma}
Consider matrix $\Psi^{\sigma}_t (x_t)$ defined as $$ \Psi^\sigma_{t}(x_t)          =\begin{bmatrix}
    \sigma_{1,t}(x_{1,t}) & \cdots & \sigma_{N,t}(x_{1,t})\\
    \vdots & \ddots & \vdots \\
    \sigma_{1,t}(x_{N,t}) & \cdots & \sigma_{N,t}(x_{N,t})
    \end{bmatrix}, $$
where $\sigma_{i,t}(x_{j,t})$ is the posterior standard deviation of agent $i$'s function evaluated at $x_{j,t}$.
Then $\sum_{t=1}^{T} \mathbf{1}_N^{\top} \Psi^{ \sigma}_{t-1} (x_t) \mathbf{1}_N \leq N^2\sqrt{4T \lambda \gamma_T}$.
\label{lemma2}
\end{lemma}

\smallskip


\noindent  \textbf{Proof of Lemma \ref{lemma2}.}
We observe that summation
\begin{align*}
    \sum_{t=1}^{T} \mathbf{1}_N^{\top} \Psi^{ \sigma}_{t-1} (x_t) \mathbf{1}_N 
    &= \sum_{t=1}^{T} \sum_{i=1}^{N} \sum_{j=1}^{N} \sigma_{i,t-1} (x_{j,t}), \\
    & \leq \sum_{j=1}^{N} \sum_{i=1}^{N} \sqrt{T \sum_{t=1}^{T} \sigma^2_{i,t-1}(x_{j,t})}, 
\end{align*} 
where the last inequality is obtained using the Cauchy-Schwartz inequality. Using the fact that $0\leq \sigma^2_{i,t-1}(x_{j,t}) \leq 1 \; \forall \; x_{j,t} \in D$, for any $ \lambda\geq 0$ we have $\lambda^{-1}\sigma^2_{j,t-1}(x_{j,t}) \leq 2\ln(1+\lambda^{-1}\sigma^2_{j,t-1}(x_{j,t}))$. This results in

\vspace{-5pt}
\begin{small}
\begin{align*}
    \sum_{t=1}^{T} \mathbf{1}_N^{\top} \Psi^{ \sigma}_{t-1} (x_t) \mathbf{1}_N &\leq \sum_{i=1}^{N} \sum_{j=1}^{N} \sqrt{2T \sum_{t=1}^{T} \lambda\ln(1+\lambda^{-1}\sigma^2_{i,t-1}(x_{j,t}))},
\end{align*}
\end{small}

Maximum information gain can be obtained as $\gamma_T = \max_{x\in \mathcal{D}}\frac{1}{2}\sum_{t=1}^{T}\ln(1+\lambda^{-1}\sigma_{i,t-1}^2(x))$ \cite{srinivas2009gaussian}. Using this we get
\begin{align*}
    \sum_{t=1}^{T} \mathbf{1}_N^{\top} \Psi^{ \sigma}_{t-1} (x_t) \mathbf{1}_N 
    &\leq \sum_{i=1}^{N} \sum_{j=1}^{N} \sqrt{4T \lambda \gamma_T}, \\ 
    &\leq N^2\sqrt{4T \lambda \gamma_T}.   
\end{align*}
This completes the proof. \hfill $\qed$

\medskip


\noindent \textbf{Proof of Theorem 1. (MA-IGP-UCB)}
We define the following vector functions:
\begin{align*}
    \overline \mu_{t}(x) = \col\{\overline \mu_{1,t}(x), \dots,\overline \mu_{N,t}(x)\},\\
    \mu_{t}(x) = \col\{\mu_{1,t}(x), \dots,\mu_{N,t}(x)\}, \\
    \overline \sigma_{t}(x) = \col\{\overline \sigma_{1,t}(x), \dots,\overline \sigma_{N,t}(x)\},\\
    \sigma_{t}(x) = \col\{\sigma_{1,t}(x), \dots,\sigma_{N,t}(x)\}.
\end{align*}
With this, the update law \eqref{eq:all_update} can be compactly written as-
\begin{subequations}
\begin{align*}
    &\overline \mu_{t}(x) =  P \overline \mu_{t-1}(x) + (\mu_{t}(x) - \mu_{t-1}(x)), \\
    &\overline \sigma_{t}(x) =  P \overline \sigma_{t-1}(x) + (\sigma_{t}(x) - \sigma_{t-1}(x)),
\end{align*}
\end{subequations}
where $P$ is the Perron matrix $(P= I-L)$, where $L$ is the Laplacian of graph $\mathcal{G}$. By unrolling the equations backward in time we get
\begin{subequations}
\begin{align}
    &\overline \mu_{t}(x) = \sum_{\tau=1}^{t-1} (P^{t-\tau}-P^{t-\tau-1}) \mu_{\tau}(x) + \mu_{t}(x), 
    \label{eq_thm1:mu_vector}\\
    &\overline \sigma_{t}(x) = \sum_{\tau=1}^{t-1} (P^{t-\tau}-P^{t-\tau-1}) \sigma_{\tau}(x) + \sigma_{t}(x),
    \label{eq_thm1:sigma_vector}
\end{align}
\end{subequations}
for $t>1$, and for $t=1$, we have $\overline \mu_{1}(x) = \mu_{1}(x)$ and $\overline \sigma_{1}(x) =\sigma_{1}(x) $. 
The sampling point, decided by agent $i$ at time step $t$ is obtained using
\begin{align*}
    x_{i,t} &= \argmax  \left(\overline \mu_{i,t-1}(x) + \beta_t \overline \sigma_{i,t-1}(x)\right). 
\end{align*}
Because of the maximum operation, the upper confidence bound evaluated at the sampling point $x_{i,t}$ exceeds the confidence bound evaluated at the actual maximum value of the global function. Therefore, for agent $i$, we obtain 
\begin{align}
     \overline \mu_{i,t-1}(x_{i,t}) + \beta_t \overline \sigma_{i,t-1}(x_{i,t}) \geq  \overline \mu_{i,t-1}(x^*) + \beta_t \overline \sigma_{i,t-1}(x^*).
     \label{eq_thm1:max_relation}
\end{align}
Before moving further, we introduce some definitions. We define $x_t = \col\{ x_{1,t}, \dots,x_{N,t}\}$. With a little abuse of notation we define  $\mu_{t}(x_t) = \col\{\mu_{1,t}(x_{1,t}), \dots,\mu_{N,t}(x_{N,t})\}$, and $ \sigma_{t}(x_t) = \col\{\sigma_{1,t}(x_{1,t}), \dots,\sigma_{N,t}(x_{N,t})\}$. Since each term has two sets of indices (in the function and in the action), we define the following matrices
\begin{align*}
    &\Psi^\mu_{t}(x_t) =\begin{bmatrix}
    \mu_{1,t}(x_{1,t}) & \cdots & \mu_{N,t}(x_{1,t})\\
    \vdots & \ddots & \vdots \\
    \mu_{1,t}(x_{N,t}) & \cdots & \mu_{N,t}(x_{N,t})
    \end{bmatrix}, \\
    &\Psi^\sigma_{t}(x_t) =\begin{bmatrix}
    \sigma_{1,t}(x_{1,t}) & \cdots & \sigma_{N,t}(x_{1,t})\\
    \vdots & \ddots & \vdots \\
    \sigma_{1,t}(x_{N,t}) & \cdots & \sigma_{N,t}(x_{N,t})
    \end{bmatrix}, \\
    &\Psi_t(x_t) = \Psi^\mu_{t}(x_t) + \beta_t\Psi^\sigma_{t}(x_t).
\end{align*}

Summing up \eqref{eq_thm1:max_relation} $\forall x =1,\dots,N$, gives us
\begin{align}
    \sum_{i=1}^{N} \Big(\overline \mu_{i,t-1}(x_{i,t}) + &\beta_t \overline \sigma_{i,t-1}(x_{i,t})\Big) \geq \nonumber\\ &
    \sum_{i=1}^{N} \left(\overline \mu_{i,t-1}(x^*) + \beta_t \overline \sigma_{i,t-1}(x^*)\right).
    \label{eq_thm1:adding_max}
\end{align}
The left-hand side of this equation can be re-written as
\begin{align*}
    \mathbf{1}_N^{\top}\begin{bmatrix}
    \overline \mu_{1,t-1}(x_{1,t}) + \beta_t \overline \sigma_{1,t-1}(x_{1,t}) \\
    \vdots \\
    \overline \mu_{N,t-1}(x_{N,t}) + \beta_t \overline \sigma_{N,t-1}(x_{N,t})
    \end{bmatrix}. 
\end{align*}

\noindent
Next, we use \eqref{eq_thm1:mu_vector} and \eqref{eq_thm1:sigma_vector} to simplify this further. We introduce the notation $[M]_{k}$ to denote the $k$th row of the matrix M. Using this, the left-hand side of the \eqref{eq_thm1:adding_max} can be simplified as

\begin{small}
\begin{align*}
    & \mathbf{1}_N^{\top}\begin{bmatrix}
    \hspace{-1.cm}
    \sum_{\tau=1}^{t-2} [P^{t-\tau-1}-P^{t-\tau-2}]_1 \mu_{\tau}(x_{1,t}) + \mu_{1,t-1}(x_{1,t})  \\  \hspace{0.1cm}
    + \beta_t (\sum_{\tau=1}^{t-2} [P^{t-\tau-1}-P^{t-\tau-2}]_{1} \sigma_{\tau}(x_{1,t}) + \sigma_{1,t-1}(x_{1,t})) \\
    \vdots \\ \hspace{-1cm}
    \sum_{\tau=1}^{t-2} [P^{t-\tau-1}-P^{t-\tau-2}]_N \mu_{\tau}(x_{N,t}) + \mu_{N,t-1}(x_{N,t})  \\  \hspace{0.1cm}
    + \beta_t (\sum_{\tau=1}^{t-2} [P^{t-\tau-1}-P^{t-\tau-2}]_N \sigma_{\tau}(x_{N,t}) + \sigma_{N,t-1}(x_{N,t}))
    \end{bmatrix} \\
    =& 
    \mathbf{1}_N^{\top}\begin{bmatrix} \hspace{-1.25cm}
    (\sum_{\tau=1}^{t-2} [P^{t-\tau-1}-P^{t-\tau-2}]_1 \odot \mu_{\tau}(x_{1,t})^{\top})\mathbf{1}_N 
    \\ 
    + \mu_{1,t-1}(x_{1,t})  \\  \hspace{0.5cm}
    + \beta_t ((\sum_{\tau=1}^{t-2} [P^{t-\tau-1}-P^{t-\tau-2}]_1 \odot\sigma_{\tau}(x_{1,t})^{\top})\mathbf{1}_N \\
    + \sigma_{1,t-1}(x_{1,t})) \\
    \vdots \\ \hspace{-1.25cm}
    (\sum_{\tau=1}^{t-2} [P^{t-\tau-1}-P^{t-\tau-2}]_N \odot \mu_{\tau}(x_{N,t})^{\top})\mathbf{1}_N 
    \\
    + \mu_{N,t-1}(x_{N,t})  \\  \hspace{0.5cm}
    + \beta_t ((\sum_{\tau=1}^{t-2} [P^{t-\tau-1}-P^{t-\tau-2}]_N \odot\sigma_{\tau}(x_{N,t})^{\top})\mathbf{1}_N \\
    + \sigma_{N,t-1}(x_{N,t})) \end{bmatrix}.
\end{align*}
\end{small}

Using the definition of $\Psi^\mu_{t}(x_t)$ and $\Psi^\sigma_{t}(x_t)$, the equation above can be compactly written as 

\begin{small}
\begin{align*}
    \mathbf{1}_N^{\top}& \bigg((\sum_{\tau=1}^{t-2} (P^{t-\tau-1}-P^{t-\tau-2}) \odot \Psi^{\mu}_\tau (x_t) )\mathbf{1}_N + \mu_{t-1}(x_t) \\
     &+ \beta_t(\sum_{\tau=1}^{t-2} (P^{t-\tau-1}-P^{t-\tau-2}) \odot \Psi^{\sigma}_\tau (x_t))\mathbf{1}_N + \beta_t\sigma_{t-1}(x_t) \bigg), \\
    = & \mathbf{1}_N^{\top} \left(\sum_{\tau=1}^{t-2} (P^{t-\tau-1}-P^{t-\tau-2}) \odot \Psi^{\mu}_\tau (x_t) \right)\mathbf{1}_N  \\ &+ 
     \mathbf{1}_N^{\top} \left(I\odot \Psi^{\mu}_{t-1} (x_t)\right)\mathbf{1}_N + \beta_t\mathbf{1}_N^{\top} \left(I\odot \Psi^{\sigma}_{t-1} (x_t)\right)\mathbf{1}_N,  \\
     & + 
     \beta_t\left(\sum_{\tau=1}^{t-2} (P^{t-\tau-1}-P^{t-\tau-2}) \odot \Psi^{\sigma}_\tau (x_t)\right)\mathbf{1}_N 
      \\
    = & \sum_{\tau=1}^{t-2} \mathbf{1}_N^{\top}\left((P^{t-\tau-1} - P^{t-\tau-2}) \odot \Psi_\tau (x_t)\right) \mathbf{1}_N \\ &+  \mathbf{1}_N^{\top}\left( I \odot\Psi_{t-1}(x_t)\right) \mathbf{1}_N,
\end{align*}
\end{small}

\noindent
which holds for $t\geq 2$, and will be $\mathbf{1}_N^{\top}\left( I \odot\Psi_{t-1}(x_t)\right) \mathbf{1}_N$ otherwise.

\noindent
One can similarly simplify the right-hand side of \eqref{eq_thm1:adding_max}, resulting in
\begin{align*}
    &\sum_{\tau=1}^{t-2}  \mathbf{1}_N^{\top} \left((P^{t-\tau-1} - P^{t-\tau-2}) \odot \Psi_\tau (x_t)\right) \mathbf{1}_N \\& \quad+  \mathbf{1}_N^{\top}\left( I \odot\Psi_{t-1}(x_t)\right) \mathbf{1}_N  \geq \\& \sum_{\tau=1}^{t-2} \mathbf{1}_N^{\top}\left((P^{t-\tau-1}-P^{t-\tau-2}) \odot \Psi_\tau (x^*)\right) \mathbf{1}_N \\ 
    & \quad+  \mathbf{1}_N^{\top}\left( I \odot \Psi_{t-1}(x^*)\right) \mathbf{1}_N. 
\end{align*}

Next, we define two matrices consisting of unknown reward functions as
\begin{align*}
    \Psi(x_t) &= \begin{bmatrix}
    f_{1}(x_{1,t}) & \cdots & f_{N}(x_{1,t})\\
    \vdots & \ddots & \vdots \\
    f_{1}(x_{N,t}) & \cdots & f_{N}(x_{N,t})
    \end{bmatrix},\\
    \Psi(x) &= \begin{bmatrix}
    f_{1}(x) & \cdots & f_{N}(x)\\
    \vdots & \ddots & \vdots \\
    f_{1}(x) & \cdots & f_{N}(x)
    \end{bmatrix}.
\end{align*}
From Lemma \ref{lemma1} it is established that $\mu_{i,t}(x) + \beta_t\sigma_{i,t}(x) \geq f_i(x)$, using which we get $\Psi^{\mu}_\tau(x_t) + \beta_t \Psi^{\sigma}_\tau(x_t) \geq \Psi(x_t)$ (these inequalities of matrices are element-wise). Hence with a probability of $1-\delta$, one can bound
\begin{align}
    &\sum_{\tau=1}^{t-2} \mathbf{1}_N^{\top}\left((P^{t-\tau-1} - P^{t-\tau-2}) \odot \Psi_\tau (x_t)\right) \mathbf{1}_N \nonumber\\
    & \quad\quad +  \mathbf{1}_N^{\top}\left( I \odot\Psi_{t-1}(x_t)\right) \mathbf{1}_N  \nonumber \geq  \\
    &  \sum_{\tau=1}^{t-2} \mathbf{1}_N^{\top}\left((P^{t-\tau-1}-P^{t-\tau-2}) \odot \Psi (x^*)\right) \mathbf{1}_N \nonumber\\
    & \quad\quad+  \mathbf{1}_N^{\top}(I \odot \Psi(x^*)) \mathbf{1}_N  \nonumber\\
    & = \mathbf{1}_N^{\top} \left((P^{t-2} - P^{t-3}) + \dots (P^{1} - I ) + I\right) \odot \Psi (x^*) \mathbf{1}_N \nonumber \\
    & = \mathbf{1}_N^{\top} P^{t-2} \odot \Psi (x^*) \mathbf{1}_N  \nonumber \\
    &  = \sum_{i=1}^{N} f_i(x^*)  \label{eq_thm1:f_star_bound},
\end{align}
where the last equation is obtained using the fact that $P$ is a doubly stochastic matrix owing to the metropolis weights \cite{xiao2006distributed}. We define scaled instantaneous pseudo-regret $r_\mathcal{G}(t) = Nr(t)$, where $r(t)$ is the  instantaneous pseudo-regret of the network. Using the definition of $r(t)$, we get
\begin{align}
    r_\mathcal{G}(t) & = \sum_{i=1}^N f_i(x^{*}) - \sum_{i=1}^N \frac{1}{N}\left(\sum_{j=1}^N f_i(x_{j,t})\right), \nonumber
\end{align}
where summation $\sum_{j=1}^N f_i(x_{j,t})$ in the last term can be written as $\mathbf{1}_{N}^{\top} \Psi(x_t) \mathbf{1}_{N}$, resulting in
\begin{align}
    r_\mathcal{G}(t) & = \sum_{i=1}^N f_i(x^{*}) - \frac{1}{N} \mathbf{1}_{N}^{\top} \Psi(x_t) \mathbf{1}_{N}.
    \label{eq:thm1_scaled_ins_regret}
\end{align}
Using the relation established in \eqref{eq_thm1:f_star_bound}, we get
\begin{align}
    r_\mathcal{G}(t) \leq & \sum_{\tau=1}^{t-2} \mathbf{1}_N^{\top}\left(P^{t-\tau-1} - P^{t-\tau-2}\right) \odot \Psi_\tau (x_t) \mathbf{1}_N \nonumber\\& +  \mathbf{1}_N^{\top}(I \odot \Psi_{t-1}(x_t))\mathbf{1}_N - \frac{1}{N} \mathbf{1}_{N}^{\top} \Psi(x_t) \mathbf{1}_{N}.
\end{align}
From lemma \ref{lemma1}, it is established that $\mu_{i,t}(x) - \beta_t\sigma_{i,t}(x) \leq f_i(x)$, using which we get $\Psi^{\mu}_\tau(x_t) - \beta_t \Psi^{\sigma}_\tau(x_t) \leq \Psi(x_t)$ or $\Psi^{\mu}_\tau(x_t) \leq \beta_t \Psi^{\sigma}_\tau(x_t) + \Psi(x_t)$ (these inequalities of matrices are element-wise). The results in
\begin{align}
    r_\mathcal{G}(t) \leq \;
    & 2\sum_{\tau=1}^{t-2} \beta_{\tau}\mathbf{1}_N^{\top}\left(P^{t-\tau-1} - P^{t-\tau-2}\right) \odot \Psi^{\sigma}_\tau (x_t) \mathbf{1}_N \nonumber\\
    & + 2\beta_t\mathbf{1}_N^{\top}(I \odot \Psi^{\sigma}_{t-1}(x_t)) \mathbf{1}_N \nonumber\\ 
    & + \sum_{\tau=1}^{t-2} \mathbf{1}_N^{\top}\left((P^{t-\tau-1} - P^{t-\tau-2}) \odot \Psi(x_t)\right) \mathbf{1}_N \nonumber\\
    & + \mathbf{1}_N^{\top}(I \odot \Psi(x_t)) \mathbf{1}_N - \frac{1}{N} \mathbf{1}_{N}^{\top} \Psi(x_t) \mathbf{1}_{N}. \label{eq_thm1:inst_regret_f}
\end{align}
Using the relation
\begin{small}
\begin{align*}
    \sum_{\tau=1}^{t-2} \mathbf{1}_N^{\top}&\left((P^{t-\tau-1} - P^{t-\tau-2}) \odot \Psi(x_t)\right) \mathbf{1}_N +  \mathbf{1}_N^{\top}( I \odot \Psi(x_t)) \mathbf{1}_N 
    \\
    & = \mathbf{1}_N^{\top} \Big((P^{t-2} - P^{t-3}) + (P^{t-3} - P^{t-4})+\\
    &\quad \quad + \dots +(P^{1} - I ) + I\Big) \odot \Psi(x_t) \mathbf{1}_N, \\
    & = \mathbf{1}_N^{\top} P^{t-2}\odot \Psi(x_t) \mathbf{1}_N \nonumber,
\end{align*}
\end{small}
the inequality in \eqref{eq_thm1:inst_regret_f} can be simplified to

\begin{align}
    r_\mathcal{G}(t) \leq \;
    & 2\sum_{\tau=1}^{t-2} \beta_{\tau}\mathbf{1}_N^{\top}\left(P^{t-\tau-1} - P^{t-\tau-2}\right) \odot \Psi^{\sigma}_\tau (x_t) \mathbf{1}_N \nonumber\\
    & + 2\beta_t\mathbf{1}_N^{\top} (I \odot \Psi^{\sigma}_{t-1}(x_t)) \mathbf{1}_N \nonumber\\ 
    & + \mathbf{1}_{N}^{\top} \left(P^{t-2}-\frac{\mathbf{1}_N\mathbf{1}^{\top}_N}{N}\right)\odot \Psi(x_t) \mathbf{1}_{N},
    \\
    \leq \;
    & 2\beta_t\sum_{\tau=1}^{t-2} \mathbf{1}_N^{\top}\left(P^{t-\tau-1} - P^{t-\tau-2}\right) \odot \Psi^{\sigma}_\tau (x_t) \mathbf{1}_N \nonumber\\
    & + 2\beta_t\mathbf{1}_N^{\top} (I \odot \Psi^{\sigma}_t(x_t)) \mathbf{1}_N \nonumber\\ 
    & + \mathbf{1}_{N}^{\top} \left(P^{t-2}-\frac{\mathbf{1}_N\mathbf{1}^{\top}_N}{N}\right)\odot \Psi(x_t) \mathbf{1}_{N}.
    \label{eq_thm1:inst_regret_beta}
\end{align} 
\noindent
Note that this relation holds only for $t\geq 2$, and otherwise will become $\mathbf{1}_N^{\top}( I \odot \Psi(x_t)) \mathbf{1}_N$.
Equation \eqref{eq_thm1:inst_regret_beta} is derived from the fact that $\beta_\tau$ is a monotonically increasing function, as maximum information gain is also monotonically increasing (as stated in Lemma \ref{lemma1}). We define the scaled cumulative regret as $R_\mathcal{G}(T) = \sum_{t=1}^{T}r_\mathcal{G}(t)$ (Note that the cumulative regret defined for the network is $R(t) = \frac{1}{N}R_\mathcal{G}(T)$). We note that at time steps $t=1$ and $t=2$, poor estimates are utilized to make decisions, resulting in significant regret. To simplify our analysis, we assume that the maximum regret, $4NB$, occurs during these two-time steps. This is based on the fact that for any function in the Reproducing Kernel Hilbert Space (RKHS), its $L_2$ norm is dominated by the RKHS norm \cite{dai2014scalable}. Using \eqref{eq_thm1:inst_regret_beta} the scaled cumulative regret can be bounded as 
\begin{align}
    R_\mathcal{G}(T) \leq \;
    & 2\beta_T \sum_{t=3}^{T} \Bigg( \sum_{\tau=1}^{t-2} \mathbf{1}_N^{\top}\left(P^{t-\tau-1} - P^{t-\tau-2}\right) \odot \Psi^{\sigma}_\tau (x_t) \mathbf{1}_N \nonumber\\ 
    & + \mathbf{1}_N^{\top} (I \odot \Psi^{\sigma}_{t-1}(x_t)) \mathbf{1}_N\Bigg) \nonumber\\ 
    & + \sum_{t=3}^{T}\mathbf{1}_{N}^{\top} \left(P^{t-2}-\frac{\mathbf{1}_N\mathbf{1}^{\top}_N}{N}\right)\odot \Psi(x_t) \mathbf{1}_{N} + 4NB.
    \label{eq_thm1:cum_regret_def}
\end{align}
Next, in order to bound the first two term of \eqref{eq_thm1:cum_regret_def}, we observe 

\begin{align}
    &\sum_{t=3}^{T}\left( \sum_{\tau=1}^{t-2} \left(P^{t-\tau-1} - P^{t-\tau-2}\right) \odot \Psi^{\sigma}_\tau (x_t)\right) \nonumber\\
    & \quad \quad + \sum_{t=3}^{T} \left( I \odot \Psi^{\sigma}_{t-1}(x_t) \right) \nonumber\\
    & = \sum_{t=1}^{T-2} \left(P^{T-t-1} - P^{T-t-2}\right) \odot \sum_{\tau=1}^{t} \Psi^{\sigma}_\tau (x_{\tau+T-t}) \nonumber\\
    & \quad \quad + \sum_{t=3}^{T} I \odot \Psi^{\sigma}_{t-1}(x_t) .
    \label{eq_thm1:cum_regret_sum}
\end{align}
Next, we study \eqref{eq_thm1:cum_regret_sum} in further depth. Since $P$ is a doubly stochastic Perron matrix, it has a maximum eigenvalue of $\lambda_1=1$ with eigenvector as $\mathbf{1}_N$. All the other eigenvalues $\lambda_p, p=2,...,N$, have a magnitude less than $1$. Let $u_p, p=1,...,N$, denote the normalized eigenvectors corresponding to eigenvalues $\lambda_p$. Hence we can write 
\begin{align*}
    P^\tau = \frac{1}{N}\mathbf{1}_N\mathbf{1}_N^{\top} + \sum_{p=2}^{N} \lambda_p^\tau u_p u_p^{\top}.
\end{align*}
Using this, the second term of \eqref{eq_thm1:cum_regret_sum} can be written as
\vspace{-5pt}

\begin{small}
\begin{align*}
    &\sum_{t=1}^{T-2} \left(\sum_{p=2}^{N} \lambda_p^{T-t-1} u_p u_p^{\top} - \sum_{p=2}^{N} \lambda_p^{T-t-2} u_p u_p^{\top}\right) \odot \sum_{\tau=1}^{t} \Psi^{\sigma}_\tau (x_{\tau+T-t}) \\
    &=\sum_{t=1}^{T-2} \left(\sum_{p=2}^{N} \lambda_p^{T-t-2} u_p u_p^{\top} (\lambda_p-1)\right) \odot \sum_{\tau=1}^{t} \Psi^{\sigma}_\tau (x_{\tau+T-t}). 
\end{align*}
\end{small}

\vspace{-5pt}
\noindent
If we consider the $ij$-th term of this matrix. we get
\begin{align*}
    &\sum_{t=1}^{T-2} \left(\sum_{p=2}^{N} \lambda_p^{T-t-2} u_p^i u_p^{j} (\lambda_p-1)\right) \sum_{\tau=1}^{t} \sigma_{i,\tau} (x_{\tau+T-t}^j) \\
    & \leq \sum_{t=1}^{T-2} \left(\sum_{p=2}^{N} |\lambda_p|^{T-t-2}|\lambda_p-1|\right) \sum_{\tau=1}^{t} \sigma_{i,\tau} (x_{\tau+T-t}^j), \\
    & \leq \sum_{t=1}^{T-2} \left(2(N-1)|\lambda_2|^{T-t-2}\right) \sum_{\tau=1}^{t} \sigma_{i,\tau} (x_{\tau+T-t}^j). 
\end{align*}
Hence the complete matrix can be written as
\begin{align*}
    \sum_{t=1}^{T-2} \left(2(N-1)|\lambda_2|^{T-t-2}\right) \sum_{\tau=1}^{t} \Psi^{\sigma}_\tau (x_{\tau+T-t}).
\end{align*}
This allows us to bound \eqref{eq_thm1:cum_regret_sum} with
\vspace{-10pt}

\begin{small}
\begin{align*}
    &\sum_{t=3}^{T} I \odot \Psi^{\sigma}_{t-1}(x_t) + \sum_{t=1}^{T-2} \left(2(N-1)|\lambda_2|^{T-t-2}\right) \sum_{\tau=1}^{t} \Psi^{\sigma}_\tau (x_{\tau+T-t}),
\end{align*}
\end{small}

\vspace{-5pt}
\noindent
resulting in a bound on the scaled cumulative regret \eqref{eq_thm1:cum_regret_def} given by 
\begin{align}
    R_\mathcal{G}(T) \leq
    & 2\beta_T \mathbf{1}_N^{\top} \Bigg(\sum_{t=3}^{T} I \odot \Psi^{\sigma}_{t-1}(x_t) \nonumber\\ 
    & + \sum_{t=1}^{T-2} \left(2(N-1)|\lambda_2|^{T-t-2}\right) \sum_{\tau=1}^{t} \Psi^{\sigma}_\tau (x_{\tau+T-t})\Bigg)\mathbf{1}_N \nonumber\\ 
    & + \sum_{t=3}^{T}\mathbf{1}_{N}^{\top} \left(P^{t-2}-\frac{\mathbf{1}_N\mathbf{1}^{\top}_N}{N}\right)\odot \Psi(x_t) \mathbf{1}_{N} + 4NB.
    \label{eq_thm1:cum_regret_final}
\end{align}
By invoking Lemma \ref{lemma2}, we bound the first term of \eqref{eq_thm1:cum_regret_final} by
\begin{align*}
    &\mathbf{1}_N^{\top} \Bigg(\sum_{t=3}^{T} I \odot \Psi^{\sigma}_{t-1}(x_t) \\& \quad \quad + \sum_{t=1}^{T-2} \left(2(N-1)|\lambda_2|^{T-t-2}\right) \sum_{\tau=1}^{t} \Psi^{\sigma}_\tau (x_{\tau+T-t})\Bigg)\mathbf{1}_N \\
    & \leq N\sqrt{4T\lambda\gamma_T} + 2(N-1)N^2\sum_{t=1}^{T-2} \left(|\lambda_2|^{T-t-2}\right) \sqrt{4t\lambda\gamma_{t}}, \\
    & \leq N\sqrt{4T\lambda\gamma_T} + 2(N-1)N^2\sum_{t=1}^{T-2} \left(|\lambda_2|^{T-t-2}\right) \sqrt{4T\lambda\gamma_T}, \\
    & \leq N\sqrt{4T\lambda\gamma_T} + 2(N-1)N^2 \sqrt{4T\lambda\gamma_T}\sum_{t=1}^{T-2} \left(|\lambda_2|^{T-t-2}\right), \\
    & \leq \left(N+ \frac{2(N-1)N^2}{1-|\lambda_2|}\right) \sqrt{4T\lambda\gamma_T} .
\end{align*}
The second last term of \eqref{eq_thm1:cum_regret_final} can be bounded as
\begin{align*}
    \sum_{t=3}^{T} \mathbf{1}_{N}^{\top} &\left(P^{t-2}-\frac{\mathbf{1}_N\mathbf{1}^{\top}_N}{N}\right)\odot \Psi(x_t) \mathbf{1}_{N} \\ 
    & = \sum_{t=3}^{T} \mathbf{1}_{N}^{\top} \sum_{p=2}^{N} \left(\lambda_p^{t-2}u_pu_p^{\top}\right)\odot \Psi(x_t) \mathbf{1}_{N}\\
    & \leq N^2(N-1)B\sum_{t=3}^{T}|\lambda_2|^{t-2} \\
    & \leq \frac{N^2(N-1)B|\lambda_2|}{1-|\lambda_2|}
    \end{align*} 
Hence we get the scaled cumulative pseudo-regret bound as
\begin{align}
    R_\mathcal{G}(T) \leq & 2\beta_T \left(N+ \frac{2(N-1)N^2}{1-|\lambda_2|}\right) \sqrt{4T\lambda\gamma_T} \nonumber \\ &
    +  \frac{N^2(N-1)B|\lambda_2|}{1-|\lambda_2|} + 4NB.
\end{align}
This completes the proof. \hfill $\qed$


\medskip

\noindent \textbf{Proof of Corollary 1}
In the case of a completely connected graph, the Perron matrix has the largest eigenvalue as $1$, and the rest of them are zero. We get $P = \frac{1}{N} \mathbf{1}_N \mathbf{1}_N^{\top}$. The update law, similar to \eqref{eq_thm1:mu_vector} and \eqref{eq_thm1:sigma_vector} can be expressed as
\begin{align*}
    &\overline \mu_{t}(x) = (P-I) \mu_{t-1}(x) + \mu_{t}(x), \\
    &\overline \sigma_{t}(x) = (P-I) \sigma_{t-1}(x) + \sigma_{t}(x).
\end{align*}
Using a similar analysis as in theorem 1, the scaled cumulative regret can be bounded as
\begin{align*}
    R_\mathcal{G}(T) \leq \;
    & 2\beta_T \mathbf{1}_N^{\top} \Bigg(\sum_{t=3}^{T} I \odot \Psi^{\sigma}_{t-1}(x_t) \\& + (P-I)\odot \sum_{t=1}^{T-2} \Psi^{\sigma}_{t-1} (x_t)\Bigg)\mathbf{1}_N + 4NB.
\end{align*}
We note that $P-I$, has diagonal terms as $\frac{1-N}{N}$ and off-diagonal terms as $\frac{1}{N}$. This results in regret bound as
\begin{align}
    R_\mathcal{G}(T) \leq & 4N\beta_T \sqrt{4T\lambda\gamma_T} +4NB. \nonumber 
\end{align}
This completes the proof. \hfill $\qed$

\medskip

\noindent \textbf{Proof of Theorem 2. (MAD-IGP-UCB)}
The proof for Theorem 2 closely follows that of Theorem 1. We will continue with the notation used in the proof for Theorem 1. The updates in MAD-IGP-UCB take place in stages of $c$ iterations. For a current stage $s$, let $t_s = (s-1)c$ denote the last time step of the previous stage. Similar to the proof of Theorem 1, we define the vector functions:
\begin{align*}
    \mu_{t}(x) &= \col\{\mu_{1,t}(x), \dots,\mu_{N,t}(x)\}, \\
    \sigma_{t}(x) &= \col\{ \sigma_{1,t}(x), \dots,\sigma_{N,t}(x)\},\\
    \overline \mu_{md, t}(x) &= \col\{\overline \mu_{md,1,t}(x), \dots,\overline \mu_{md,N,t}(x)\}, \\
    \overline \sigma_{md,t}(x)  &= \col\{\overline \sigma_{md,1,t}(x), \dots,\overline \sigma_{md,N,t}(x)\}.
\end{align*}
The mixed estimates for the current stage get fixed at $t_s$ and can be written as
\vspace{-10pt}

\begin{small}
\begin{subequations}
\begin{align}
    &\overline \mu_{md,t_s}(x) = \sum_{\tau=1}^{t_s-c-1} (P^{t_s-\tau}-P^{t_s-\tau-1}) \mu_{\tau}(x) + P^{c} \mu_{t_s-c}(x),  
    \label{eq_th2:mu_expansion}\\
    &\overline \sigma_{md,t_s}(x) = \sum_{\tau=1}^{t_s-c-1} (P^{t_s-\tau}-P^{t_s-\tau-1}) \sigma_{\tau}(x)+ P^{c} \sigma_{t_s-c}(x).  
    \label{eq_th2:sigma_expansion}
\end{align}
\end{subequations}
\end{small}

\vspace{-10pt}
\noindent
The sampling point decided by agent $i$ at time step $t$, is obtained using
\begin{align*}
    x_{i,t} &= \argmax  \left(\overline \mu_{md,i,ts}(x) + \beta_{t_s} \overline \sigma_{md,i,ts}(x)\right). 
\end{align*}
Using the maximum operation, the upper confidence bound evaluated at the sampling point $x^i_t$ exceeds the confidence bound evaluated at the actual maximum value of the global function and hence, we have
\begin{align}
     \overline \mu_{md,i,ts}(x_{i,t}) + \beta_t \overline \sigma_{md,i,ts}(x_{i,t}) \geq  \overline \mu_{md,i,ts}(x^*) + \beta_t \overline \sigma_{md,i,ts}(x^*).
     \label{eq_th2:max_relation}
\end{align}

Since each term on the left hand side of the above equation has two sets of indices (in the function and the action), with a slight abuse of notation we define the following matrices
\begin{align*}
    &\Psi^{\mu_{md}}_{t}(x_t) =\begin{bmatrix}
    \mu_{md,1,t}(x_{1,t}) & \cdots & \mu_{md,N,t}(x_{1,t})\\
    \vdots & \ddots & \vdots \\
    \mu_{md,1,t}(x_{N,t}) & \cdots & \mu_{md,N,t}(x_{N,t})
    \end{bmatrix},\\
    &\Psi^{\sigma_{md}}_{t}(x_t) =\begin{bmatrix}
    \sigma_{md,1,t}(x_{1,t}) & \cdots & \sigma_{md,N,t}(x_{1,t})\\
    \vdots & \ddots & \vdots \\
    \sigma_{md,1,t}(x_{N,t}) & \cdots & \sigma_{md,N,t}(x_{N,t})
    \end{bmatrix}, \\
    &\Psi_t(x_t) = \Psi^{\mu_{md}}_{t}(x_t) + \beta_t\Psi^{\sigma_{md}}_{t}(x_t).
\end{align*}

Using the similar procedure considered in the proof of Theorem 1, summation of \eqref{eq_th2:max_relation} $\forall \; i =1,\dots,N$ and using \eqref{eq_th2:mu_expansion} and \eqref{eq_th2:sigma_expansion} gives us
\begin{align}
    \sum_{\tau=1}^{t_s-c-1} &\mathbf{1}_N^{\top}\left((P^{t_s-\tau} - P^{t_s-\tau-1}) \odot \Psi_\tau (x_t)\right) \mathbf{1}_N \nonumber\\ 
    & \quad +  \mathbf{1}_N^{\top}\left(P^{c}\odot \Psi_{t_s-c} (x_t) \right)\mathbf{1}_N   \nonumber\\ 
    & \geq \sum_{\tau=1}^{t_s-c-1} \mathbf{1}_N^{\top}\left((P^{t_s-\tau}-P^{t_s-\tau-1}) \odot \Psi_\tau (x^*)\right) \mathbf{1}_N \nonumber\\ 
    & \quad +  \mathbf{1}_N^{\top}\left( P^{c} \odot \Psi_{t_s-c}(x^*)\right) \mathbf{1}_N, \nonumber\\
    & \geq \sum_{\tau=1}^{t_s-c-1} \mathbf{1}_N^{\top}\left((P^{t_s-\tau}-P^{t_s-\tau-1}) \odot \Psi (x^*)\right) \mathbf{1}_N \nonumber\\ 
    & \quad +  \mathbf{1}_N^{\top}(P^{c} \odot \Psi(x^*)) \mathbf{1}_N,  \label{eq_th2:max_inequality}\\
    & = \mathbf{1}_N^{\top} \Big((P^{t_s-1} - P^{t_s-2})  \nonumber\\ 
    & \quad  + \dots +(P^{c+1} - P^{c} ) + P^{c} \Big) \odot \Psi (x^*) \mathbf{1}_N, \nonumber \\
    & = \mathbf{1}_N^{\top} P^{t_s-1} \odot \Psi (x^*) \mathbf{1}_N,  \nonumber \\
    &  = \sum_{i=1}^{N} f_i(x^*)  \label{eq_th2:reg_f_star}, 
\end{align}
where \eqref{eq_th2:max_inequality} was achieved using Lemma \ref{lemma1} and the last equations comes from the fact, that $P$ is doubly stochastic matrix. Recall the definition of scaled instantaneous pseudo-regret $r_\mathcal{G}(t) = Nr(t)$ and relation \eqref{eq:thm1_scaled_ins_regret} used in proof of Theorem 1. Using the relation established in \eqref{eq_th2:reg_f_star}, $r_\mathcal{G}(t)$ can be bounded as
\begin{align*}
    r_\mathcal{G}(t) \leq & \sum_{\tau=1}^{t_s-c-1} \mathbf{1}_N^{\top}\left(P^{t_s-\tau} - P^{t_s-\tau-1}\right) \odot \Psi_\tau (x_t) \mathbf{1}_N \\ 
    & +  \mathbf{1}_N^{\top}(P^{c} \odot \Psi_{t_s-c}(x_t))\mathbf{1}_N - \frac{1}{N} \mathbf{1}_{N}^{\top} \Psi(x_t) \mathbf{1}_{N}. 
\end{align*}
From lemma \ref{lemma1}, it is established that $\mu_{i,t}(x) - \beta_t\sigma_{i,t}(x) \leq f_i(x)$ and so will be $\mu_{md,i,t}(x) - \beta_t\sigma_{md,i,t}(x) \leq f_i(x)$, using which we get $\Psi^{\mu_{md}}_\tau(x_t) - \beta_t \Psi^{\sigma_{md}}_\tau(x_t) \leq \Psi(x_t)$ or $\Psi^{\mu_{md}}_\tau(x_t) \leq \beta_t \Psi^{\sigma_{md}}_\tau(x_t) + \Psi(x_t)$ (these inequalities of matrices are element-wise). This results in
\begin{align}
    r_\mathcal{G}(t) \leq \;
    & 2\sum_{\tau=1}^{t_s-c-1} \beta_{\tau}\mathbf{1}_N^{\top}\left(P^{t_s-\tau} - P^{t_s-\tau-1}\right) \odot \Psi^{\sigma_{md}}_\tau (x_t) \mathbf{1}_N \nonumber\\ 
    & +  2\beta_{t_s-c}\mathbf{1}_N^{\top}(P^{c} \odot \Psi^{\sigma_{md}}_{t_s-c}(x_t)) \mathbf{1}_N \nonumber\\ 
    &+\sum_{\tau=1}^{t_s-c-1} \mathbf{1}_N^{\top}\left((P^{t_s-\tau} - P^{t_s-\tau-1}) \odot \Psi(x_t)\right) \mathbf{1}_N \nonumber\\ 
    & +  \mathbf{1}_N^{\top}(P^{c}  \odot \Psi(x_t)) \mathbf{1}_N - \frac{1}{N} \mathbf{1}_{N}^{\top} \Psi(x_t) \mathbf{1}_{N}. \label{eq_th2:inst_regret_sigma2}
\end{align}
Using this fact that
\begin{align*}
    &\sum_{\tau=1}^{t_s-c-1} \mathbf{1}_N^{\top}\left((P^{t_s-\tau} - P^{t_s-\tau-1}) \odot \Psi(x_t)\right) \mathbf{1}_N \nonumber\\ 
    & \quad \quad+  \mathbf{1}_N^{\top}( P^{c} \odot \Psi(x_t)) \mathbf{1}_N \\
    & = \mathbf{1}_N^{\top} \Big((P^{t_s-1} - P^{t_s-2}) \nonumber\\ 
    & \quad \quad+\dots +(P^{c+1} - P^{c} ) + P^{c} \Big) \odot \Psi(x_t) \mathbf{1}_N \\
    & = \mathbf{1}_N^{\top} P^{t_s-1}\odot \Psi(x_t) \mathbf{1}_N,
\end{align*}
where this holds for $t_s > c+1$, hence for $t>2c$, the inequality in \eqref{eq_th2:inst_regret_sigma2} can be simplified to
\begin{align}
    r_\mathcal{G}(t) \leq \;
    & 2\sum_{\tau=1}^{t_s-c-1} \beta_{\tau}\mathbf{1}_N^{\top}\left(P^{t_s-\tau} - P^{t_s-\tau-1}\right) \odot \Psi^{\sigma_{md}}_\tau (x_t) \mathbf{1}_N \nonumber\\ 
    & +  2\beta_{t_s-c}\mathbf{1}_N^{\top} (P^{c} \odot \Psi^{\sigma_{md}}_{t_s-c}(x_t)) \mathbf{1}_N \nonumber\\ 
    & + \mathbf{1}_{N}^{\top} \left(P^{t_s-1}-\frac{\mathbf{1}_N\mathbf{1}^{\top}_N}{N}\right)\odot \Psi(x_t) \mathbf{1}_{N},
    \nonumber\\
    \leq \;
    & 2\beta_{t_s-c}\sum_{\tau=1}^{t_s-c-1} \mathbf{1}_N^{\top}\left(P^{t_s-\tau} - P^{t_s-\tau-1}\right) \odot \Psi^{\sigma_{md}}_\tau (x_t) \mathbf{1}_N \nonumber\\ 
    & +  2\beta_{t_s-c}\mathbf{1}_N^{\top} (P^{c} \odot \Psi^{\sigma_{md}}_{t_s-c}(x_t)) \mathbf{1}_N \nonumber\\ 
    & + \mathbf{1}_{N}^{\top} \left(P^{t_s-1}-\frac{\mathbf{1}_N\mathbf{1}^{\top}_N}{N}\right)\odot \Psi(x_t) \mathbf{1}_{N}.
    \label{eq_th2:inst_regret_beta}
\end{align}
Where \eqref{eq_th2:inst_regret_beta} is obtained using the fact that $\beta_\tau$ is a monotonically increasing function. We define scaled cumulative regret as $R_\mathcal{G}(T) = \sum_{t=1}^{T}r_\mathcal{G}(t)$ (Note that cumulative regret defined for the network is $R(t) = \frac{1}{N}R_\mathcal{G}(T)$). We note that until stage $s=3$, i.e. for $t\leq 2c$, no mixed estimates are available to make decisions. We use maximum regret that occurred for these time steps resulting in a regret of $4cNB$, using the fact that for any function in RKHS, its $L_2$ norm is dominated by the RKHS norm \cite{dai2014scalable}. Using \eqref{eq_th2:inst_regret_beta} the scaled cumulative regret can be bounded as

\begin{small}
\begin{align}
    R_\mathcal{G}(T) \leq
    & 2\beta_T \sum_{t=2c+1}^{T} \Bigg( \sum_{\tau=1}^{t_s-c-1} \mathbf{1}_N^{\top}\left(P^{t_s-\tau} - P^{t_s-\tau-1}\right) \odot \Psi^{\sigma_{md}}_\tau (x_t) \mathbf{1}_N \nonumber\\ 
    & + \mathbf{1}_N^{\top} (P^{c} \odot \Psi^{\sigma_{md}}_{t_s-c}(x_t)) \mathbf{1}_N\Bigg) \nonumber\\ 
    & + \sum_{t=2c+1}^{T}\mathbf{1}_{N}^{\top} \left(P^{t_s-1}-\frac{\mathbf{1}_N\mathbf{1}^{\top}_N}{N}\right)\odot \Psi_B \mathbf{1}_{N} + 4cNB.
    \label{eq_th2:cum_regret_def}
\end{align}
\end{small}

Next, in order to bound the first term of \eqref{eq_th2:cum_regret_def}, we utilize the fact that for $t$ in any particular stage $s$, the actions taken by the agent will remain fixed and hence the network will incur a constant regret during that stage. Hence the first term of \eqref{eq_th2:cum_regret_def} can be re-written as
\begin{align}
    &\sum_{t=2c+1}^{T} \Bigg(\sum_{\tau=1}^{t_s-c-1} \left(P^{t_s-\tau} - P^{t_s-\tau-1}\right) \odot \Psi^{\sigma_{md}}_\tau (x_t) \nonumber\\ 
    & \quad \quad + P^{c} \odot \Psi^{\sigma_{md}}_{t_s-c}(x_t) \Bigg) \nonumber\\
    & = \sum_{ t=2c+1}^{T} P^{c} \odot \Psi^{\sigma_{md}}_{t_s-c}(x_t) \nonumber\\ 
    & \quad \quad + \sum_{t=2c+1}^{ T}\sum_{\tau=1}^{t_s-c-1} \left(P^{t_s-\tau} - P^{t_s-\tau-1}\right) \odot \Psi^{\sigma_{md}}_\tau (x_t) \nonumber\\
    & = \sum_{ t=2c+1}^{T} P^{c} \odot \Psi^{\sigma_{md}}_{t_s-c}(x_t) + c\sum_{t=1}^{T-2c-1}  \Big(P^c(P^{T-2c-t} \nonumber\\ 
    & \quad \quad- P^{T-2c-t-1})\Big) \odot \sum_{\tau=1}^{\llfloor \frac{t}{c} \rrfloor}\Psi^{\sigma_{md}}_{\lb \tau -\llfloor \frac{t}{c} \rrfloor \rb c+t} (x_k),\label{eq:thm2_r22}
\end{align} 
where $k=\lb \tau - \llfloor \frac{t}{c} \rrfloor + \llfloor \frac{T}{c} \rrfloor\rb(c-1)+1$.
\noindent
Next, we study the second term of \eqref{eq:thm2_r22} further depth. Given $P$ is a doubly stochastic Perron matrix, it has a maximum eigenvalue of $\lambda_1=1$ with eigenvector as $\textbf{1}_N$. All the other eigenvalues $\lambda_p, p=2,...,n$ has a magnitude less than $1$. Let $u_p, p=1,...,n$ denote the normalized eigenvectors corresponding to eigenvalues $\lambda_p$. Hence we can write 
\begin{align*}
    P^\tau = \frac{1}{N}\mathbf{1}_N\mathbf{1}_N + \sum_{p=2}^{N} \lambda_p^\tau u_p u_p^{\top}.
\end{align*}
Using this, we get
\begin{small}
\begin{align}
    &c\sum_{t=1}^{T-2c-1} P^c \left(P^{T-2c-t} - P^{T-2c-t-1}\right) \odot \sum_{\tau=1}^{\llfloor \frac{t}{c} \rrfloor}\Psi^{\sigma_{md}}_{\lb \tau -\llfloor \frac{t}{c} \rrfloor \rb c+t} (x_k)\nonumber\\
    & = c\hspace{-4pt}\sum_{t=1}^{T-2c-1} \hspace{-3pt}\left(\sum_{p=2}^{N} \lambda_p^{T-c-t-1} u_p u_p^{\top} (\lambda_p-1)\right) \odot \sum_{\tau=1}^{\llfloor \frac{t}{c} \rrfloor}\Psi^{\sigma_{md}}_{\lb \tau -\llfloor \frac{t}{c} \rrfloor \rb c+t} (x_k) \nonumber
\end{align}
\end{small}

If we consider the $ij$th term of this matrix we get
\begin{small}
\begin{align*}
    &c\sum_{t=1}^{T-2c-1} \left(\sum_{p=2}^{N} \lambda_p^{T-c-t-1} u_p^i u_p^{j} (\lambda_p-1)\right) \sum_{\tau=1}^{\llfloor \frac{t}{c} \rrfloor}  \sigma_{md,i,\lb \tau -\llfloor \frac{t}{c} \rrfloor \rb c+t} (x_k) \\
    & \leq 2c(N-1) |\lambda_2|^{c} \hspace{-2pt} \sum_{t=1}^{T-2c-1} \hspace{-4pt} \left(|\lambda_2|^{T-2c-t-1}\right)\sum_{\tau=1}^{\llfloor \frac{t}{c} \rrfloor}  \sigma_{md,i,\lb \tau -\llfloor \frac{t}{c} \rrfloor \rb c+t} (x_k)
\end{align*}
\end{small}

The complete matrix can be written as
\begin{align}
    2c(N-1)|\lambda_2|^{c}\sum_{t=1}^{T-2c-1} \left(|\lambda_2|^{T-2c-t-1}\right)  \sum_{\tau=1}^{\llfloor \frac{t}{c} \rrfloor}\Psi^{\sigma_{md}}_{\lb \tau -\llfloor \frac{t}{c} \rrfloor \rb c+t} (x_k).
\end{align}
This gives us the scaled cumulative regret \eqref{eq_th2:cum_regret_def} as

\begin{small}
\begin{align}
    R&_\G(T) \leq
     2\beta_T \mathbf{1}_N^{\top} \Bigg(\sum_{t=2c+1}^{T} P^{c}  \odot \Psi^{\sigma_{md}}_{t_s-c}(x_t) \nonumber\\ 
    & + 2c(N-1)|\lambda_2|^{c} \sum_{t=1}^{T-2c-1} |\lambda_2|^{T-2c-t-1}\sum_{\tau=1}^{\llfloor \frac{t}{c} \rrfloor}\Psi^{\sigma_{md}}_{\lb \tau -\llfloor \frac{t}{c} \rrfloor \rb c+t} (x_k)\Bigg)\mathbf{1}_N \nonumber\\ 
    & + \sum_{t=2c+1}^{T}\mathbf{1}_{N}^{\top} \left(P^{t_s-1}-\frac{\mathbf{1}_N\mathbf{1}^{\top}_N}{N}\right)\odot \Psi(x_t) \mathbf{1}_{N} + 4cNB.
    \label{eq_th2:cum_regret_n}
\end{align}
\end{small}

\noindent
By invoking Lemma \ref{lemma2}, we bound the first term of \eqref{eq_th2:cum_regret_n} by
\begin{align*}
    \mathbf{1}_N^{\top} &\Bigg(\sum_{t=2c+1}^{T} P^{c}  \odot \Psi^{\sigma_{md}}_{t_s-c}(x_t) + 2(N-1)|\lambda_2|^{c} \\& \quad \sum_{t=1}^{T-2c-1} |\lambda_2|^{T-2c-t-1}c\sum_{\tau=1}^{\llfloor \frac{t}{c} \rrfloor}\Psi^{\sigma_{md}}_{\lb \tau -\llfloor \frac{t}{c} \rrfloor \rb c+t} (x_k)\Bigg)\mathbf{1}_N \\
    & \leq N^2\sqrt{4T\lambda\gamma_T} \\& \quad + 2(N-1)N^2 |\lambda_2|^{c}\sum_{t=1}^{T-2c-1} \left(|\lambda_2|^{T-2c-t-1}\right) \sqrt{4t\lambda\gamma_t} \\
    & \leq N^2\sqrt{4T\lambda\gamma_T} \\& \quad + 2(N-1)N^2|\lambda_2|^{c}\sum_{t=1}^{T-2c-1} \left(|\lambda_2|^{T-2c-t-1}\right) \sqrt{4T\lambda\gamma_T} \\
    & \leq N^2\sqrt{4T\lambda\gamma_T} \\& \quad + 2(N-1)N^2 |\lambda_2|^{c}\sqrt{2T\sigma^2\gamma_T}\sum_{t=1}^{T-2c-1} \left(|\lambda_2|^{T-2c-t-1}\right) \\
    & \leq \left(N^2+ \frac{2(N-1)N^2|\lambda_2|^{c}}{1-|\lambda_2|}\right) \sqrt{4T\lambda\gamma_T} 
\end{align*}
Hence giving us the scaled cumulative regret bound as
\begin{align}
    R_\G(T) \leq
    & 2\beta_T \left(N^2+ \frac{2(N-1)N^2|\lambda_2|^{c}}{1-|\lambda_2|}\right) \sqrt{2T\sigma^2\gamma_T} \nonumber 
    \nonumber\\ 
    & + \sum_{t=2c+1}^{T}\mathbf{1}_{N}^{\top} \left(P^{t_s-1}-\frac{\mathbf{1}_N\mathbf{1}^{\top}_N}{N}\right)\odot \Psi_B \mathbf{1}_{N} + 4cNB.
    \label{eq_th2:cum_regret6}
\end{align}
Now the second last term of \eqref{eq_th2:cum_regret6}, can be bounded as
\begin{align*}
    \sum_{t=2c+1}^{T}& \mathbf{1}_{N}^{\top} \left(P^{t_s-1}-\frac{\mathbf{1}_N\mathbf{1}^{\top}_N}{N}\right)\odot \Psi(x_t) \mathbf{1}_{N} \\
    & = \sum_{t=2c+1}^{T} \mathbf{1}_{N}^{\top} \sum_{p=2}^{N} \left(\lambda_p^{t_s-1}u_pu_p^{\top}\right)\odot \Psi(x_t) \mathbf{1}_{N}, \\
    & \leq N^2(N-1)B\sum_{t=2c+1}^{T}|\lambda_2|^{t_s-1}, \\
    & \leq \frac{N^2(N-1)B|\lambda_2|^{2c}}{1-|\lambda_2|}.
    \end{align*}
\noindent
Hence we get the regret bound as
\begin{align}
    R_\mathcal{G}(T) \leq & 2\beta_T \left(N^2+ \frac{2(N-1)N^2|\lambda_2|^{c}}{1-|\lambda_2|}\right) \sqrt{4T\lambda\gamma_T} \nonumber \\&
    +  \frac{N^2(N-1)B|\lambda_2|^{2c}}{1-|\lambda_2|} + 4cNB.
\end{align}
This completes the proof. \hfill $\qed$

\section*{References}
\bibliographystyle{IEEEtran}
\bibliography{sample}

\end{document}

%% file: macros.tex
\newcommand{\A}{\mathcal{A}}

\newcommand{\R}{\mathbb{R}}
\newcommand{\oo}{\mathcal{O}}

\newcommand{\G}{\mathcal{G}}

\newcommand{\rrfloor}{\right\rfloor}
\newcommand{\llfloor}{\left\lfloor}
\newcommand{\rb}{\right)}
\newcommand{\lb}{\left(}

\def\qed{ \rule{.1in}{.1in}}

\usepackage{generic}
\usepackage{cite}
\usepackage{amsmath,amssymb,amsfonts}
\usepackage{bbm}
\usepackage{epsfig}
\usepackage{color}
\usepackage{graphicx}
\usepackage{algpseudocode}
\usepackage{algorithm}
\usepackage{caption}
\usepackage{subcaption}
\usepackage{braket}
\usepackage{xfrac}
\usepackage{hyperref}
\hypersetup{hidelinks}
\usepackage{textcomp}

\newcommand{\col}{{\rm col\;}}

\newtheorem{theorem}{Theorem}

\newtheorem{lemma}{Lemma}
\newtheorem{remark}{Remark}

\newtheorem{corollary}{Corollary}

\newtheorem{assumption}{Assumption}

\let\temp\rmdefault
\usepackage{mathpazo}
\let\rmdefault\temp

\DeclareMathOperator*{\argmax}{arg\,max}

%% file: main.bbl
\begin{thebibliography}{10}
\providecommand{\url}[1]{#1}
\csname url@samestyle\endcsname
\providecommand{\newblock}{\relax}
\providecommand{\bibinfo}[2]{#2}
\providecommand{\BIBentrySTDinterwordspacing}{\spaceskip=0pt\relax}
\providecommand{\BIBentryALTinterwordstretchfactor}{4}
\providecommand{\BIBentryALTinterwordspacing}{\spaceskip=\fontdimen2\font plus
\BIBentryALTinterwordstretchfactor\fontdimen3\font minus \fontdimen4\font\relax}
\providecommand{\BIBforeignlanguage}[2]{{%
\expandafter\ifx\csname l@#1\endcsname\relax
\typeout{** WARNING: IEEEtran.bst: No hyphenation pattern has been}%
\typeout{** loaded for the language `#1'. Using the pattern for}%
\typeout{** the default language instead.}%
\else
\language=\csname l@#1\endcsname
\fi
#2}}
\providecommand{\BIBdecl}{\relax}
\BIBdecl

\bibitem{yang2019survey}
T.~Yang, X.~Yi, J.~Wu, Y.~Yuan, D.~Wu, Z.~Meng, Y.~Hong, H.~Wang, Z.~Lin, and K.~H. Johansson, ``A survey of distributed optimization,'' \emph{Annual Reviews in Control}, vol.~47, pp. 278--305, 2019.

\bibitem{zheng2022review}
Y.~Zheng and Q.~Liu, ``A review of distributed optimization: Problems, models and algorithms,'' \emph{Neurocomputing}, vol. 483, pp. 446--459, 2022.

\bibitem{vakili2021optimal}
S.~Vakili, N.~Bouziani, S.~Jalali, A.~Bernacchia, and D.-s. Shiu, ``Optimal order simple regret for gaussian process bandits,'' \emph{Advances in Neural Information Processing Systems}, vol.~34, pp. 21\,202--21\,215, 2021.

\bibitem{janz2020bandit}
D.~Janz, D.~Burt, and J.~Gonz{\'a}lez, ``Bandit optimisation of functions in the mat{\'e}rn kernel rkhs,'' in \emph{International Conference on Artificial Intelligence and Statistics}.\hskip 1em plus 0.5em minus 0.4em\relax PMLR, 2020, pp. 2486--2495.

\bibitem{srinivas2009gaussian}
N.~Srinivas, A.~Krause, S.~M. Kakade, and M.~Seeger, ``Gaussian process optimization in the bandit setting: No regret and experimental design,'' \emph{arXiv preprint arXiv:0912.3995}, 2009.

\bibitem{cano2016towards}
I.~Cano, M.~Weimer, D.~Mahajan, C.~Curino, and G.~M. Fumarola, ``Towards geo-distributed machine learning,'' \emph{arXiv preprint arXiv:1603.09035}, 2016.

\bibitem{anandkumar2011distributed}
A.~Anandkumar, N.~Michael, A.~K. Tang, and A.~Swami, ``Distributed algorithms for learning and cognitive medium access with logarithmic regret,'' \emph{IEEE Journal on Selected Areas in Communications}, vol.~29, no.~4, pp. 731--745, 2011.

\bibitem{mcmahan2017communication}
B.~McMahan, E.~Moore, D.~Ramage, S.~Hampson, and B.~A. y~Arcas, ``Communication-efficient learning of deep networks from decentralized data,'' in \emph{Artificial intelligence and statistics}.\hskip 1em plus 0.5em minus 0.4em\relax PMLR, 2017, pp. 1273--1282.

\bibitem{tran2012long}
L.~Tran-Thanh, A.~Rogers, and N.~R. Jennings, ``Long-term information collection with energy harvesting wireless sensors: a multi-armed bandit based approach,'' \emph{Autonomous Agents and Multi-Agent Systems}, vol.~25, pp. 352--394, 2012.

\bibitem{boyd2011distributed}
S.~Boyd, N.~Parikh, E.~Chu, B.~Peleato, J.~Eckstein \emph{et~al.}, ``Distributed optimization and statistical learning via the alternating direction method of multipliers,'' \emph{Foundations and Trends{\textregistered} in Machine learning}, vol.~3, no.~1, pp. 1--122, 2011.

\bibitem{nedic2018distributed}
A.~Nedi{\'c} and J.~Liu, ``Distributed optimization for control,'' \emph{Annual Review of Control, Robotics, and Autonomous Systems}, vol.~1, pp. 77--103, 2018.

\bibitem{kanagawa2018gaussian}
M.~Kanagawa, P.~Hennig, D.~Sejdinovic, and B.~K. Sriperumbudur, ``Gaussian processes and kernel methods: A review on connections and equivalences,'' \emph{arXiv preprint arXiv:1807.02582}, 2018.

\bibitem{auer2002finite}
P.~Auer, N.~Cesa-Bianchi, and P.~Fischer, ``Finite-time analysis of the multiarmed bandit problem,'' \emph{Machine learning}, vol.~47, pp. 235--256, 2002.

\bibitem{agrawal1995continuum}
R.~Agrawal, ``The continuum-armed bandit problem,'' \emph{SIAM journal on control and optimization}, vol.~33, no.~6, pp. 1926--1951, 1995.

\bibitem{dubey2020cooperative}
A.~Dubey \emph{et~al.}, ``Cooperative multi-agent bandits with heavy tails,'' in \emph{International conference on machine learning}.\hskip 1em plus 0.5em minus 0.4em\relax PMLR, 2020, pp. 2730--2739.

\bibitem{li2022communication}
C.~Li, H.~Wang, M.~Wang, and H.~Wang, ``Communication efficient distributed learning for kernelized contextual bandits,'' \emph{arXiv preprint arXiv:2206.04835}, 2022.

\bibitem{du2021collaborative}
Y.~Du, W.~Chen, Y.~Kuroki, and L.~Huang, ``Collaborative pure exploration in kernel bandit,'' \emph{arXiv preprint arXiv:2110.15771}, 2021.

\bibitem{salgia2023collaborative}
S.~Salgia, S.~Vakili, and Q.~Zhao, ``Collaborative learning in kernel-based bandits for distributed users,'' \emph{IEEE Transactions on Signal Processing}, 2023.

\bibitem{dai2020federated_bay}
Z.~Dai, B.~K.~H. Low, and P.~Jaillet, ``Federated bayesian optimization via thompson sampling,'' \emph{Advances in Neural Information Processing Systems}, vol.~33, pp. 9687--9699, 2020.

\bibitem{nedic2009distributed}
A.~Nedic and A.~Ozdaglar, ``Distributed subgradient methods for multi-agent optimization,'' \emph{IEEE Transactions on Automatic Control}, vol.~54, no.~1, pp. 48--61, 2009.

\bibitem{lobel2010distributed}
I.~Lobel and A.~Ozdaglar, ``Distributed subgradient methods for convex optimization over random networks,'' \emph{IEEE Transactions on Automatic Control}, vol.~56, no.~6, pp. 1291--1306, 2010.

\bibitem{aybat2017distributed}
N.~S. Aybat, Z.~Wang, T.~Lin, and S.~Ma, ``Distributed linearized alternating direction method of multipliers for composite convex consensus optimization,'' \emph{IEEE Transactions on Automatic Control}, vol.~63, no.~1, pp. 5--20, 2017.

\bibitem{gharesifard2013distributed}
B.~Gharesifard and J.~Cort{\'e}s, ``Distributed continuous-time convex optimization on weight-balanced digraphs,'' \emph{IEEE Transactions on Automatic Control}, vol.~59, no.~3, pp. 781--786, 2013.

\bibitem{kia2015distributed}
S.~S. Kia, J.~Cort{\'e}s, and S.~Mart{\'\i}nez, ``Distributed convex optimization via continuous-time coordination algorithms with discrete-time communication,'' \emph{Automatica}, vol.~55, pp. 254--264, 2015.

\bibitem{jakovetic2014fast}
D.~Jakoveti{\'c}, J.~Xavier, and J.~M. Moura, ``Fast distributed gradient methods,'' \emph{IEEE Transactions on Automatic Control}, vol.~59, no.~5, pp. 1131--1146, 2014.

\bibitem{xi2018linear}
C.~Xi, V.~S. Mai, R.~Xin, E.~H. Abed, and U.~A. Khan, ``Linear convergence in optimization over directed graphs with row-stochastic matrices,'' \emph{IEEE Transactions on Automatic Control}, vol.~63, no.~10, pp. 3558--3565, 2018.

\bibitem{dhingra2018proximal}
N.~K. Dhingra, S.~Z. Khong, and M.~R. Jovanovi{\'c}, ``The proximal augmented lagrangian method for nonsmooth composite optimization,'' \emph{IEEE Transactions on Automatic Control}, vol.~64, no.~7, pp. 2861--2868, 2018.

\bibitem{aybat2015asynchronous}
N.~Aybat, Z.~Wang, and G.~Iyengar, ``An asynchronous distributed proximal gradient method for composite convex optimization,'' in \emph{International Conference on Machine Learning}.\hskip 1em plus 0.5em minus 0.4em\relax PMLR, 2015, pp. 2454--2462.

\bibitem{tatarenko2017non}
T.~Tatarenko and B.~Touri, ``Non-convex distributed optimization,'' \emph{IEEE Transactions on Automatic Control}, vol.~62, no.~8, pp. 3744--3757, 2017.

\bibitem{pang2019randomized}
Y.~Pang and G.~Hu, ``Randomized gradient-free distributed optimization methods for a multiagent system with unknown cost function,'' \emph{IEEE Transactions on Automatic Control}, vol.~65, no.~1, pp. 333--340, 2019.

\bibitem{yuan2014randomized}
D.~Yuan and D.~W. Ho, ``Randomized gradient-free method for multiagent optimization over time-varying networks,'' \emph{IEEE Transactions on Neural Networks and Learning Systems}, vol.~26, no.~6, pp. 1342--1347, 2014.

\bibitem{shahrampour2017}
S.~Shahrampour, A.~Rakhlin, and A.~Jadbabaie, ``Multi-armed bandits in multi-agent networks,'' in \emph{2017 IEEE International Conference on Acoustics, Speech and Signal Processing (ICASSP)}.\hskip 1em plus 0.5em minus 0.4em\relax IEEE, 2017, pp. 2786--2790.

\bibitem{landgren2016distributed}
P.~Landgren, V.~Srivastava, and N.~E. Leonard, ``On distributed cooperative decision-making in multiarmed bandits,'' in \emph{2016 European Control Conference (ECC)}.\hskip 1em plus 0.5em minus 0.4em\relax IEEE, 2016, pp. 243--248.

\bibitem{landgren2016distributed2}
------, ``Distributed cooperative decision-making in multiarmed bandits: Frequentist and bayesian algorithms,'' in \emph{2016 IEEE 55th Conference on Decision and Control (CDC)}.\hskip 1em plus 0.5em minus 0.4em\relax IEEE, 2016, pp. 167--172.

\bibitem{moradipari2022collaborative}
A.~Moradipari, M.~Ghavamzadeh, and M.~Alizadeh, ``Collaborative multi-agent stochastic linear bandits,'' in \emph{2022 American Control Conference (ACC)}.\hskip 1em plus 0.5em minus 0.4em\relax IEEE, 2022, pp. 2761--2766.

\bibitem{chakraborty2017coordinated}
M.~Chakraborty, K.~Y.~P. Chua, S.~Das, and B.~Juba, ``Coordinated versus decentralized exploration in multi-agent multi-armed bandits.'' in \emph{IJCAI}, 2017, pp. 164--170.

\bibitem{korda2016distributed}
N.~Korda, B.~Szorenyi, and S.~Li, ``Distributed clustering of linear bandits in peer to peer networks,'' in \emph{International conference on machine learning}.\hskip 1em plus 0.5em minus 0.4em\relax PMLR, 2016, pp. 1301--1309.

\bibitem{szorenyi2013gossip}
B.~Szorenyi, R.~Busa-Fekete, I.~Hegedus, R.~Orm{\'a}ndi, M.~Jelasity, and B.~K{\'e}gl, ``Gossip-based distributed stochastic bandit algorithms,'' in \emph{International conference on machine learning}.\hskip 1em plus 0.5em minus 0.4em\relax PMLR, 2013, pp. 19--27.

\bibitem{martinez2019decentralized}
D.~Mart{\'\i}nez-Rubio, V.~Kanade, and P.~Rebeschini, ``Decentralized cooperative stochastic bandits,'' \emph{Advances in Neural Information Processing Systems}, vol.~32, 2019.

\bibitem{scaman2017optimal}
K.~Scaman, F.~Bach, S.~Bubeck, Y.~T. Lee, and L.~Massouli{\'e}, ``Optimal algorithms for smooth and strongly convex distributed optimization in networks,'' in \emph{international conference on machine learning}.\hskip 1em plus 0.5em minus 0.4em\relax PMLR, 2017, pp. 3027--3036.

\bibitem{zhu2023distributed}
J.~Zhu and J.~Liu, ``Distributed multi-armed bandits,'' \emph{IEEE Transactions on Automatic Control}, 2023.

\bibitem{bistritz2018distributed}
I.~Bistritz and A.~Leshem, ``Distributed multi-player bandits-a game of thrones approach,'' \emph{Advances in Neural Information Processing Systems}, vol.~31, 2018.

\bibitem{mehrabian2020practical}
A.~Mehrabian, E.~Boursier, E.~Kaufmann, and V.~Perchet, ``A practical algorithm for multiplayer bandits when arm means vary among players,'' in \emph{International Conference on Artificial Intelligence and Statistics}.\hskip 1em plus 0.5em minus 0.4em\relax PMLR, 2020, pp. 1211--1221.

\bibitem{mitra2021exploiting}
A.~Mitra, H.~Hassani, and G.~Pappas, ``Exploiting heterogeneity in robust federated best-arm identification,'' \emph{arXiv preprint arXiv:2109.05700}, 2021.

\bibitem{shi2021per}
C.~Shi, C.~Shen, and J.~Yang, ``Federated multi-armed bandits with personalization,'' in \emph{International conference on artificial intelligence and statistics}.\hskip 1em plus 0.5em minus 0.4em\relax PMLR, 2021, pp. 2917--2925.

\bibitem{reda2022near}
C.~R{\'e}da, S.~Vakili, and E.~Kaufmann, ``Near-optimal collaborative learning in bandits,'' \emph{arXiv preprint arXiv:2206.00121}, 2022.

\bibitem{brochu2010tutorial}
E.~Brochu, V.~M. Cora, and N.~De~Freitas, ``A tutorial on bayesian optimization of expensive cost functions, with application to active user modeling and hierarchical reinforcement learning,'' \emph{arXiv preprint arXiv:1012.2599}, 2010.

\bibitem{chowdhury2017kernelized}
S.~R. Chowdhury and A.~Gopalan, ``On kernelized multi-armed bandits,'' in \emph{International Conference on Machine Learning}.\hskip 1em plus 0.5em minus 0.4em\relax PMLR, 2017, pp. 844--853.

\bibitem{abbasi2011improved}
Y.~Abbasi-Yadkori, D.~P{\'a}l, and C.~Szepesv{\'a}ri, ``Improved algorithms for linear stochastic bandits,'' \emph{Advances in neural information processing systems}, vol.~24, 2011.

\bibitem{valko2013finite}
M.~Valko, N.~Korda, R.~Munos, I.~Flaounas, and N.~Cristianini, ``Finite-time analysis of kernelised contextual bandits,'' \emph{arXiv preprint arXiv:1309.6869}, 2013.

\bibitem{krause2011contextual}
A.~Krause and C.~Ong, ``Contextual gaussian process bandit optimization,'' \emph{Advances in neural information processing systems}, vol.~24, 2011.

\bibitem{bogunovic2016}
I.~Bogunovic, J.~Scarlett, and V.~Cevher, ``Time-varying gaussian process bandit optimization,'' in \emph{Artificial Intelligence and Statistics}.\hskip 1em plus 0.5em minus 0.4em\relax PMLR, 2016, pp. 314--323.

\bibitem{dubey2020kernel}
A.~Dubey \emph{et~al.}, ``Kernel methods for cooperative multi-agent contextual bandits,'' in \emph{International Conference on Machine Learning}.\hskip 1em plus 0.5em minus 0.4em\relax PMLR, 2020, pp. 2740--2750.

\bibitem{lilearning}
C.~Li, H.~Wang, M.~Wang, and H.~Wang, ``Learning kernelized contextual bandits in a distributed and asynchronous environment,'' in \emph{The Eleventh International Conference on Learning Representations}, 2022.

\bibitem{shi2021federated}
C.~Shi and C.~Shen, ``Federated multi-armed bandits,'' in \emph{Proceedings of the AAAI Conference on Artificial Intelligence}, vol.~35, no.~11, 2021, pp. 9603--9611.

\bibitem{Chaloner1995}
K.~Chaloner and I.~Verdinelli, ``Bayesian experimental design: A review,'' \emph{Source: Statistical Science}, vol.~10, pp. 273--304, 1995.

\bibitem{vakili2021information}
S.~Vakili, K.~Khezeli, and V.~Picheny, ``On information gain and regret bounds in gaussian process bandits,'' in \emph{International Conference on Artificial Intelligence and Statistics}.\hskip 1em plus 0.5em minus 0.4em\relax PMLR, 2021, pp. 82--90.

\bibitem{chen2012distributed}
F.~Chen, Y.~Cao, and W.~Ren, ``Distributed average tracking of multiple time-varying reference signals with bounded derivatives,'' \emph{IEEE Transactions on Automatic Control}, vol.~57, no.~12, pp. 3169--3174, 2012.

\bibitem{scarlett2017lower}
J.~Scarlett, I.~Bogunovic, and V.~Cevher, ``Lower bounds on regret for noisy gaussian process bandit optimization,'' in \emph{Conference on Learning Theory}.\hskip 1em plus 0.5em minus 0.4em\relax PMLR, 2017, pp. 1723--1742.

\bibitem{leskovec2016snap}
J.~Leskovec and R.~Sosi{\v{c}}, ``Snap: A general-purpose network analysis and graph-mining library,'' \emph{ACM Transactions on Intelligent Systems and Technology (TIST)}, vol.~8, no.~1, pp. 1--20, 2016.

\bibitem{huang2021federated}
R.~Huang, W.~Wu, J.~Yang, and C.~Shen, ``Federated linear contextual bandits,'' \emph{Advances in neural information processing systems}, vol.~34, pp. 27\,057--27\,068, 2021.

\bibitem{dubey2020differentially}
A.~Dubey and A.~Pentland, ``Differentially-private federated linear bandits,'' \emph{Advances in Neural Information Processing Systems}, vol.~33, pp. 6003--6014, 2020.

\bibitem{xiao2006distributed}
L.~Xiao, S.~Boyd, and S.~Lall, ``Distributed average consensus with time-varying metropolis weights,'' \emph{Automatica}, vol.~1, 2006.

\bibitem{dai2014scalable}
B.~Dai, B.~Xie, N.~He, Y.~Liang, A.~Raj, M.-F.~F. Balcan, and L.~Song, ``Scalable kernel methods via doubly stochastic gradients,'' \emph{Advances in neural information processing systems}, vol.~27, 2014.

\end{thebibliography}
